\documentclass[sigconf]{acmart}
\usepackage{balance} 
\usepackage{algorithm}
\usepackage{algorithmic}

\usepackage{multirow} 
\usepackage{booktabs}
\usepackage{threeparttable}
\usepackage{subfigure} 
\usepackage{bbding} 
\usepackage{graphicx} 

\usepackage{amsmath,amsthm,amssymb,amsfonts}
\usepackage{bm}
\newcommand{\mathbi}[1]{\textbf{\em{#1}}}
\definecolor{mygreen}{RGB}{34,139,34}

\usepackage{float}

\usepackage{color}
\newtheorem{definition}{Definition}

\newtheorem{proposition}{Proposition}

\allowdisplaybreaks 
\AtBeginDocument{%
  }

\copyrightyear{2025}
\acmYear{2025}
\setcopyright{acmlicensed}
\acmConference[KDD '25]{Proceedings of the 31st ACM SIGKDD Conference on Knowledge Discovery and Data Mining V.1}{August 3--7, 2025}{Toronto, ON, Canada}
\acmBooktitle{Proceedings of the 31st ACM SIGKDD Conference on Knowledge Discovery and Data Mining V.1 (KDD '25), August 3--7, 2025, Toronto, ON, Canada}
\acmDOI{10.1145/3690624.3709226}
\acmISBN{979-8-4007-1245-6/25/08}

\begin{document}

\title{Brain Effective Connectivity Estimation via Fourier Spatiotemporal Attention}

\author{Wen Xiong}
\affiliation{%
  \institution{Beijing University of Technology}
  \state{Beijing}
  \country{China}}
\email{xiongwen@emails.bjut.edu.cn}

\author{Jinduo Liu*}
\affiliation{%
  \institution{Beijing University of Technology}
  \state{Beijing}
  \country{China}}
\email{jinduo@bjut.edu.cn}
\thanks{*Corresponding author: Jinduo Liu (email: jinduo@bjut.edu.cn)}

\author{Junzhong Ji}
\affiliation{%
  \institution{Beijing University of Technology}
  \state{Beijing}
  \country{China}}
\email{jjz01@bjut.edu.cn}

\author{Fenglong Ma}
\affiliation{%
 \institution{Pennsylvania State University}
 \country{University Park, PA, USA}}
\email{fenglong@psu.edu}





\begin{abstract}
Estimating brain effective connectivity (EC) from functional magnetic resonance imaging (fMRI) data can aid in comprehending the neural mechanisms underlying human behavior and cognition, providing a foundation for disease diagnosis.
However, current spatiotemporal attention modules handle temporal and spatial attention separately,  extracting temporal and spatial features either sequentially or in parallel. These approaches overlook the inherent spatiotemporal correlations present in real world fMRI data. Additionally, the presence of noise in fMRI data further limits the performance of existing methods.
In this paper, we propose a novel brain effective connectivity estimation method based on \underline{F}ourier \underline{s}patio\underline{t}emporal \underline{a}ttention (FSTA-EC), which combines Fourier attention and spatiotemporal attention to simultaneously capture inter-series (spatial) dynamics and intra-series (temporal) dependencies from high-noise fMRI data. 
Specifically, Fourier attention is designed to convert the high-noise fMRI data to frequency domain, and map the denoised fMRI data back to physical domain, and spatiotemporal attention is crafted to simultaneously learn spatiotemporal dynamics.
Furthermore, through a series of proofs, we demonstrate that incorporating learnable filter into fast Fourier transform and inverse fast Fourier transform processes is mathematically equivalent to performing cyclic convolution.
The experimental results on simulated and real-resting-state fMRI datasets demonstrate that the proposed method exhibits superior performance when compared to state-of-the-art methods.
The code is available at 
\url{https://github.com/XiongWenXww/FSTA}.
\end{abstract}




\maketitle

\section{Introduction}
Graph models have gained popularity as an effective approach for comprehending extensive datasets and offering outputs that are easily interpretable \cite{pamfil2020dynotears, liu2023discovering}.
Brain effective connectivity (EC) represents the influence exerted by one brain region upon another in terms of information flow and causality \cite{friston1994functional}. 
Brain EC analysis has found applications across several clinical research domains, delving into neurological and psychiatric disorders, including Alzheimer's disease \cite{Alzheimer}, schizophrenia \cite{schizophrenia}, depression \cite{depression, xinusing}, autism spectrum disorders \cite{Autism}, et al. By extrapolating brain EC insights from neuroimaging data, such as functional magnetic resonance imaging (fMRI), researchers can potentially establish biomarkers for early detection, disease progression monitoring, and treatment efficacy assessment. This also enables us to illuminate the interplay among distinct cerebral areas and their contributions to diverse cognitive processes and behaviors.

Lately, plenty of techniques have emerged that leverage fMRI data for the analysis of brain EC. These techniques can be broadly categorized into two main categories: model-based methods and data-driven methods \cite{review}.
Model-based methods are relatively straightforward to implement and offer explanatory insights into brain activity, which can reveal specific relationships and mechanisms between brain regions but require prior knowledge \cite{gilson2020model}. These methods could be constrained by inaccurate prior knowledge and assumptions incorporated during modeling \cite{farruggia2022functional}.
On the other hand, data-driven methods do not rely on priori assumptions and can adapt to different types of data. 
However, fMRI time series data exhibit strong temporal and spatial characteristics, and current data-driven methods always capture the spatial and temporal correlations separately. Thus, they disregard the unified \textbf{spatiotemporal interdependence} inherent in real-world fMRI time series data. In addition, the \textbf{high noise} levels in fMRI data also limit the performance of existing methods. Therefore, how to capture the unified spatiotemporal interdependence inherent in fMRI data and improve the model's resistance to noise have become key research challenges in this field.


Attention mechanisms \cite{attention} have been widely used in fMRI data analysis and have achieved good performance \cite{antonello2023scaling}. 
However, when the noise level in the data is high, its performance is usually affected \cite{lin2016neural}. Current studies show that employing the fast Fourier transform (FFT) and inverse fast Fourier transform (IFFT) can efficiently mitigate the effects of noise in data \cite{tao2006research}. Additionally, combining them with a novel spatiotemporal attention mechanism may also be expected to capture the unified spatiotemporal interdependence inherent in high-noise fMRI data.

In this paper, we propose a novel method that utilizes Fourier spatiotemporal attention (FSTA-EC) for brain effective connectivity estimation. FSTA-EC integrates Fourier attention (FA) and spatiotemporal attention (STA) to enable the concurrent learning of inter-series (spatial) dynamics and intra-series (temporal) dependencies from high-noise fMRI data. In detail, FA exploits FFT to transform high-noised fMRI data into  frequency domain and employs IFFT to restore the denoised fMRI data back to the original physical domain, while extracting the global frequency domain features of fMRI data during this procedure. Then, STA utilizes temporal attention (TA) to learn intricate temporal patterns from fMRI time series data and employs spatiotemporal fusion attention (STFA) to simultaneously capture both spatiotemporal features and spatial brain effective connectivity among brain regions. The proposed method has been rigorously tested with both simulated and real-world fMRI data. The experimental results demonstrate that it offers performance advantages over existing state-of-the-art methods.

The main contributions of this paper are summarized:
\begin{itemize}
\item To the best of our knowledge, this is the first work that takes into account Fourier transformation for denoising fMRI time series data and employs spatiotemporal attention to estimating brain effective connectivity.
\item The proposed method employs a learnable filter in FFT and IFFT techniques to denoise the fMRI data and extract global frequency domain features, thereby enhancing the performance on high-noise fMRI data.
\item The proposed method derives both spatiotemporal features and spatial effective connectivity from fMRI data by a unified spatiotemporal attention framework, enabling a more comprehensive utilization of the spatiotemporal relationships.
\item Systematic experiments demonstrate that in the majority of cases, the performance of FSTA-EC is superior to current state-of-the-art methods.
\end{itemize}

\section{Related Work}


\subsection{Brain Effective Connectivity Methods}
Recently, a plethora of techniques have surfaced to estimate brain EC using fMRI data. These techniques can be broadly categorized into two main groups: model-based methods and data-driven methods  \cite{review}. 
A multitude of model-based methods have been developed to estimate EC. These methods include dynamic causal modeling (DCM) \cite{DCM,spDCM}, structural equation modeling (SEM) \cite{SEM}, and Granger causality (GC)\cite{lsGC}. 
Specifically, GC is proficient at deducing causal relationships between variables; however, it remains incapable of ascertaining the directionality of these relationships and necessitates stationary data. In the case of DCM and SEM, they both offer avenues for addressing non-stationary fMRI data. But all of these model-based approaches rely on prior knowledge, which can be prone to errors.
Data-driven methods such as pairwise linear non-Gaussian acyclic model (pwLiNGAM) \cite{pw}, generative adversarial networks (GAN) \cite{EC-GAN,MCAN}, ant colony optimization based on conditional entropy and transfer entropy (ACOCTE) \cite{ACOCTE}, causal recurrent variational autoencoder (CR-VAE) \cite{CR-VAE} and causal autoencoder with meta-knowledge (MetaCAE) \cite{MetaCAE}, aiming to discover EC patterns directly from fMRI data without prior knowledge. Nevertheless, the majority of methods tend to omit the unified spatiotemporal correlation and inherent noise present in fMRI data.

\subsection{Fourier Attention}
Several studies have explored applications based on Fourier transform. Nguyen et al. employed a generalized Fourier nonparametric regression estimator while retaining QKV components \cite{nguyen2022fourierformer}. 
Patro et al. apply a normalization layer before FFT in the spectral block, and use a weighted gating mechanism after FFT, employing a common product operation \cite{patro2023spectformer}.
Lee-Thorp et al. perform discrete Fourier transform along the hidden dimension and the sequence dimension, respectively \cite{lee2021fnet}.
Unlike the aforementioned work, we incorporate an element-wise multiplication into the FFT and IFFT, which functions as a filter to control the gain of each frequency component.

\subsection{Spatiotemporal Attention}
Regarding spatiotemporal attention, many spatiotemporal attention modules have been applied to tasks such as traffic prediction, functional connectivity \cite{huang2022spatio, he2023multi} and disease classification tasks \cite{nguyen2020attend, delfan2024hybrid}. Lan et al. feed traffic data into temporal attention, followed by passing the extracted temporal features into spatial attention \cite{lan2022dstagnn}. Additionally, many studies treat temporal and spatial attention as independent modules and arrange them in a sequential structure. \cite{zhao2017video, fu2019spatiotemporal, jiang2023characterizing}. In contrast to previous studies, Zhang et al. apply temporal and spatial attention in parallel, ultimately fusing these features through gated fusion \cite{zhang2023traffic}.
However, these studies organize temporal and spatial attention either serially or in parallel,  which can only extract temporal and spatial features of fMRI data in separate patterns. These approaches fail to account for inherent spatiotemporal correlations present in real world fMRI data. In contrast, our spatiotemporal attention module fuses temporal and spatial attention within a unified framework, achieving a deep integration of temporal features and spatial relationships among nodes.


\section{Notation and Problem Definition}
In this paper, we utilize a lowercase font (e.g., $v_i$) to denote the ($i$-th) brain region, the uppercase font (e.g., $X$) represents the matrix, and the uppercase calligraphic letters (e.g., $\mathcal{X}$) are used to denote three-dimensional tensor. Let a tensor $\mathcal{X}\in\mathbb{R}^{T\times N\times D}$ signifies embedded fMRI time series data in three-dimensional, where $T$ is the number of data points, $N$ signifies the number of brain regions, and $D$ stands for the embedding size of fMRI data. Then $\mathcal{X}_t$ implies the data of the tensor $\mathcal{X}$ at time step $t$. The math bold italic letters at the subscript position of the parameter, i.e., $Q_{\mathbi{T},h}$, $Q_{\mathbi{S},h}$ denote query in temporal and spatiotemporal fusion attention mechanism, respectively. The detailed meanings of all symbols are shown in Appendix ~\ref{apd_notation}.

First, we give the definition of brain effective connectivity. 
\begin{definition}
(\textbf{Brain Effective Connectivity}) Brain effective connectivity, also known as the causal brain network, can be represented as a directed graph $\mathcal{G} = <v, A>$, where $v$ denotes the set of nodes, with each node $v_i\in v$ indicating a brain region or a region of interest (ROI). $A$ represents edge matrix, which means $A_{ij}$ signifies the effective connectivity from brain region $v_j$ to $v_i$, implying a causal influence exerted by brain region $v_j$ on $v_i$.
\end{definition}

Then, the problem of brain effective connectivity estimation can be formulated as learning $\mathcal{G}$ from brain data (e.g., fMRI).

\section{Methodology}\label{Methodology}
In this section, we first present the main idea of FSTA-EC, then describe the details of each component, and finally introduce the loss function.

\subsection{Model Overview}
Over the past few years, numerous attention-based methods have emerged for the estimation of brain EC. However, these methods have shown limited effectiveness in handling the high noise and frequency domain features of fMRI data. 
Moreover, the majority of spatiotemporal attention methods use separate patterns, inherently challenging the unified spatiotemporal interdependence observed in the real world, leading to a significant impact on brain EC estimation.

\begin{figure*}[t]
\begin{center}
\includegraphics[width=0.80\textwidth]{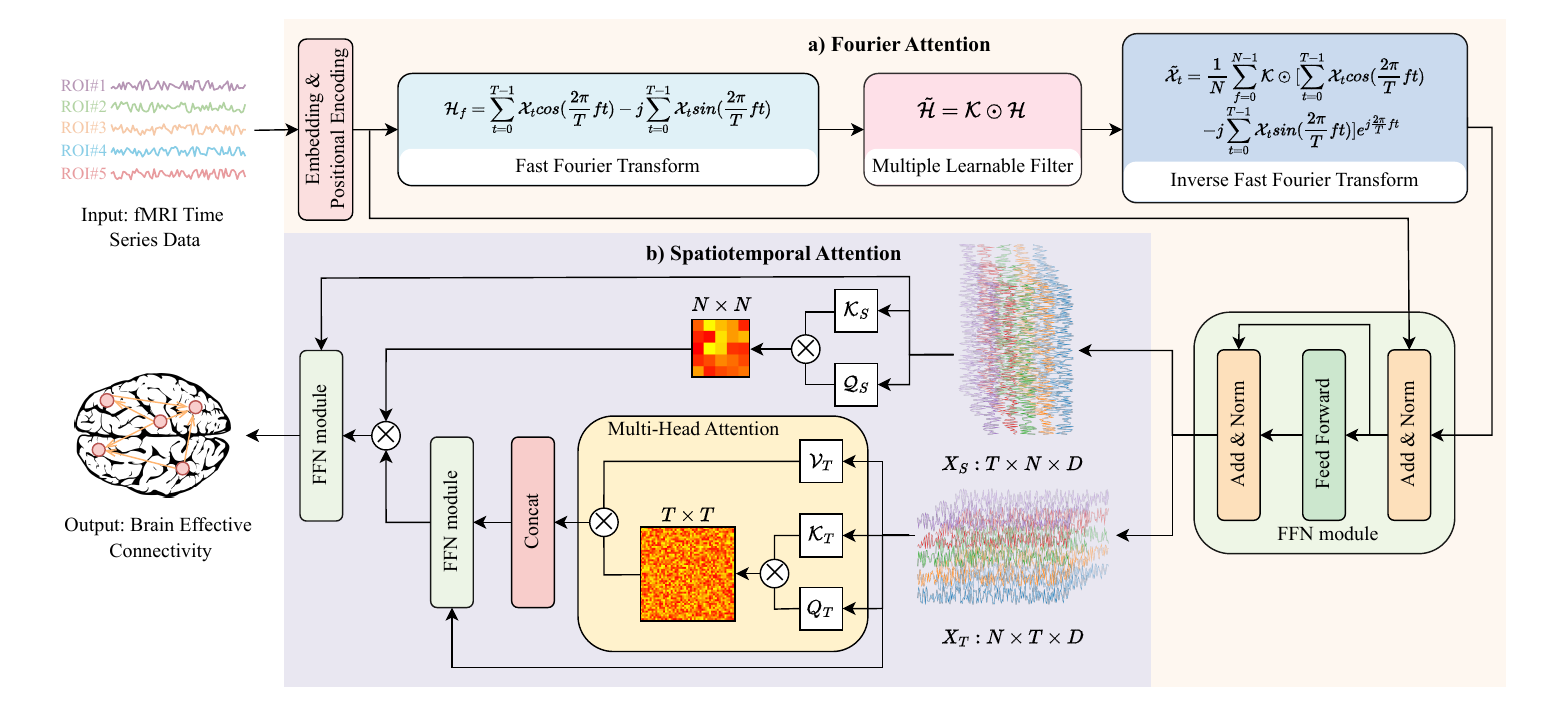}
\end{center}
\caption{The architecture of FSTA-EC. Specifically, it comprises two elements: (a) Fourier Attention and (b) Spatiotemporal Attention. Fourier attention initially applies fast Fourier transform to convert high-noise fMRI data into frequency domain and then utilizes inverse fast Fourier transform to map the denoised fMRI data back to physical domain. Spatiotemporal attention incorporates temporal attention to learn temporal features from fMRI time series data and employs spatiotemporal fusion attention to simultaneously capture both spatiotemporal dynamic features and spatial brain EC. 
}
\label{architecture}
\end{figure*}

To overcome the above issues, we present a novel brain effective connectivity estimation method via Fourier spatiotemporal attention, named FSTA-EC. The framework of FSTA-EC is depicted in Figure ~\ref{architecture}. Firstly, the fMRI data is fed into FA to remove noise and capture global context features from the frequency domain. Subsequently, the temporal dependencies are initially acquired using temporal attention in spatiotemporal attention. Then the spatiotemporal dynamic features of fMRI data and brain EC are obtained through spatiotemporal fusion attention.

\subsection{Fourier Attention}
The key difference between Fourier attention (FA) and general attention is the replacement of multi-head attention with Fourier transform. 
Firstly, to unveil the intricate features and patterns that may be present in the fMRI data, each node undergoes an embedding process at each time step, yielding three-dimensional fMRI data with dimensions $T \times N \times D$, where $T$ is the number of data points, $N$ signifies the number of nodes, and $D$ stands for the embedding size.
Due to the inability of attention to comprehend the time order of the input fMRI time series data and positional encoding (PE) has the advantage of capturing the relative position information of the sequences,  PE is needed to make the position information self-contained in the input fMRI time series data:
\begin{equation}
\begin{gathered}
PE_{(p,2k)}=\sin\left(\frac{p}{\gamma^{\frac{2k}{D}}}\right), 
PE_{(p,2k+1)}=\cos\left(\frac{p}{\gamma^{\frac{2k}{D}}}\right), \\
\mathcal{X} = (\otimes X) + PE,
\label{PE}
\end{gathered}
\end{equation}
where $p$ denotes the position of the vector in the fMRI time series data, $k$ signifies the position in the vector, 
$\gamma$ normally takes the value of 10000, and $\otimes$ represents two-dimensional convolution with the kernel size of 1 × 1.

Subsequently, the embedded fMRI time series data is passed through fast Fourier transform (FFT) for further analysis. This transformation decomposes fMRI time series data into its constituent frequencies, enabling the extraction of crucial frequency features and aiding in the identification of periodic or trending patterns within the fMRI data. Considering this, we map the embedded fMRI data to frequency domain $\mathcal{H}$ by FFT:
\begin{equation}
\mathcal{H}_f=\mathcal{F}(\mathcal{X}_t)=\sum_{t=0}^{T-1}\mathcal{X}_t e^{-j \frac{2 \pi}{T} f t},
\end{equation}
where $\mathcal{F}(\cdot)$ implies one-dimensional fast Fourier transform, $j$ is the imaginary unit, $\mathcal{X}_t$ stands for the data corresponding to the $t$-th time step of the fMRI time series, and $\mathcal{H}_f$ represents the spectrum of the sequence $\mathcal{X}_t$ at frequency $\frac{2\pi f}{T}$. According to the Euler's formula $e^{ix}=\cos x+j\sin x$ \cite{Euler}, we can further elaborate on the aforementioned equation:
\begin{equation}
\begin{aligned}
\mathcal{H}_f
=& \sum_{t=0}^{T-1}\mathcal{X}_t\cos(-\frac{2 \pi}{T} ft)+j\sum_{t=0}^{T-1}\mathcal{X}_t\sin(-\frac{2 \pi}{T} ft) \\
=&\sum_{t=0}^{T-1}\mathcal{X}_t\cos(\frac{2 \pi}{T} ft)-j\sum_{t=0}^{T-1}\mathcal{X}_t\sin(\frac{2 \pi}{T} ft).
\label{FFT}
\end{aligned}
\end{equation}

Then FA modifies the spectrum by applying a learnable filter $\mathcal{K}$ through the process of modulation:
\begin{equation}
\tilde{\mathcal{H}}=\mathcal{K}\odot \mathcal{H},
\label{filter}
\end{equation}
where $\odot$ is element-wise multiplication, and $\mathcal{K}$ is a learnable filter. 
After completing the learning process in frequency domain, we recover the original fMRI data by IFFT, which also facilitates noise reduction of the fMRI data. Similar to FFT, we can derive the following formula using Euler's formula:
\begin{equation}
\begin{aligned}
\tilde{\mathcal{X}}_t
= \mathcal{F}^{-1}(\tilde{\mathcal{H}}_f) 
=&\frac{1}{N}\sum_{f=0}^{N-1}\tilde{\mathcal{H}}_fe^{j \frac{2 \pi}{T} f t} \\
=&\frac{1}{N}\sum_{f=0}^{N-1}\mathcal{K}\odot \Bigl[\sum_{t=0}^{T-1}\mathcal{X}_t\cos(\frac{2 \pi}{T} ft)\\
&-j\sum_{t=0}^{T-1}\mathcal{X}_t\sin(\frac{2 \pi}{T} ft) \Bigr]e^{j\frac{2\pi}{T}ft},
\label{IFFT}
\end{aligned}
\end{equation}
where $\mathcal{F}^{-1}(\cdot)$ signifies one-dimensional inverse fast Fourier transform. Indeed, the spectrum is composed of cosine and sine waves with varying frequencies and amplitudes in $\mathcal{H}$. This enables us to deduce the specific periodic properties of the fMRI time series signal. Consequently, analyzing the spectrum facilitates improved identification of dominant frequencies and recurring patterns within the fMRI time series.

\begin{proposition}
\label{proposition}
By incorporating learnable filter $\mathcal{K}$ into the FFT and IFFT processes, the resulting operation is equivalent to cyclic convolution.
\end{proposition}
The detailed proof is shown in Appendix ~\ref{apd_proof}. Using a learnable filter $\mathcal{K}$ and applying it as a cyclic convolution, FA can capture long-range dependencies or patterns extending beyond the immediate neighbors. This allows FA to have a broader receptive field and consider the global context when making predictions or extracting features from the fMRI time series data.

Once the fMRI data is denoised, to prevent gradient vanishing and exploding, we incorporate residual connections and layer normalization techniques:
\begin{equation}
\tilde{\mathcal{X}}'=\rho(\tilde{\mathcal{X}}+\mathcal{X}),
\label{ffn1}
\end{equation}
where $\rho(\cdot)$ denotes layer normalization operation.
To enhance the capability of FA in capturing intricate representations, we incorporate a feedforward network (FFN) module:
\begin{equation}
\mathcal{X}'=\rho(ReLU(\tilde{\mathcal{X}}'W_1)W_2+\tilde{\mathcal{X}}').
\label{ffn2}
\end{equation}

We define $\Phi(\cdot)$ as the representation of a sequence of operations within the FFN, specifically to combine Eqs. (~\ref{ffn1}) and (~\ref{ffn2}) into:
\begin{equation}
\mathcal{X}'=\Phi(\tilde{\mathcal{X}}).
\end{equation}


\subsection{Spatiotemporal Attention}
Unlike most other spatiotemporal feature extractors, spatiotemporal attention (STA) can simultaneously extract spatiotemporal correlations and effective connectivity between different brain regions.
This is achieved through the utilization of temporal attention and spatiotemporal fusion attention.
In addition, STA benefits from multi-head attention, with the ability to simultaneously gather information from distinct subspaces, thereby enhancing the capacity to capture diverse facets of the input sequence.
Specifically, the multi-head attention is composed of multiple self-attention. In this context, we employ $H$ self-attention mechanisms. Within each temporal attention, $\mathcal{Q}_{{\mathbi{T}},h}$, $\mathcal{K}_{{\mathbi{T}},h}$, $\mathcal{V}_{{\mathbi{T}},h}$ is first obtained through linear transformation: 
\begin{equation}
\begin{aligned}
\mathcal{Q}_{{\mathbi{T}},h}=\mathcal{X}'^{\top}_{h}W_{\mathbi{T}}^Q+bias_{{\mathbi{T}},h}^Q, \\
\mathcal{K}_{{\mathbi{T}},h}=\mathcal{X}'^{\top}_{h}W_{\mathbi{T}}^K+bias_{{\mathbi{T}},h}^K, \\
\mathcal{V}_{{\mathbi{T}},h}=\mathcal{X}'^{\top}_{h}W_{\mathbi{T}}^V+bias_{{\mathbi{T}},h}^V,
\end{aligned}
\end{equation}
where $\mathcal{X}'^{\top}_{h}$ denotes the transpose of input of the $h$-th head after embedding,
$\mathcal{X}'^{\top}_{h}\in\mathbb{R}^{N\times T \times \frac{D}{H}}$, $W_{\mathbi{T}}^Q\in\mathbb{R}^{\frac{D}{H} \times \frac{D}{H}}$, $\mathcal{Q}_{{\mathbi{T}},h}$, $\mathcal{K}_{{\mathbi{T}},h}$, $\mathcal{V}_{{\mathbi{T}},h}$ symbolize the Query, Key, Value of the $h$-th head in temporal attention, respectively. Then temporal features can be calculated as:
 
\begin{equation}
\mathcal{O}_{{\mathbi{T}},h}=softmax\left(\frac{\mathcal{Q}_{{\mathbi{T}},h}\mathcal{K}_{{\mathbi{T}},h}^{\top}}{\sqrt{d}}\right)\mathcal{V}_{{\mathbi{T}},h},
\end{equation}
where $d$ denotes the dimension of the Query vector. In our specific case, $d$ is equal to $\frac{D}{H}$, so the total computation of multi-head attention is unchanged. By concatenating the information obtained from each head and then performing a linear transformation, we obtain the output of temporal attention:
\begin{equation}
\mathcal{O}_{\mathbi{T}}=\gamma(\mathcal{O}_{{\mathbi{T}},1},\mathcal{O}_{{\mathbi{T}},2},...,\mathcal{O}_{{\mathbi{T}},H})W_{\mathbi{C}},
\end{equation}
where $\gamma(\cdot)$ is the concatenation operation. After obtaining the concatenation $\mathcal{O}_{\mathbi{T}}$, similar to the approach employed in FA, the temporal features of the fMRI data are derived using the FFN module:
\begin{equation}
\mathcal{Z}_{\mathbi{T}}=\Phi(\mathcal{O}_{\mathbi{T}}).
\label{getZ_T}
\end{equation}

Given the interconnections and inherent spatiotemporal correlations among brain regions, it becomes crucial to concurrently acquire spatiotemporal features.
Therefore, in spatiotemporal fusion attention, for the output denoised fMRI data $\mathcal{X}'$ of FA with dimensions $T\times N\times D$, after acquiring $\mathcal{Q}_{{\mathbi{S}},h}$, $\mathcal{K}_{{\mathbi{S}},h}$ through linear transformation, the corresponding attention weight and brain EC can be obtained by following formula:
\begin{equation}
\begin{gathered}
\mathcal{E}_h=softmax(\frac{\mathcal{Q}_{{\mathbi{S}},h}\mathcal{K}_{{\mathbi{S}},h}^{\top}}{\sqrt{d}}),\\
\mathbf{A} = \frac{1}{T}\sum_{t=1}^T\frac{1}{H}\sum_{h=1}^H\mathcal{E}_{h,t}, 
\end{gathered}   
\end{equation}
where $\mathcal{E}_h$ indicates the attention weight of the $h$-th head, 
$\mathbf{A}$ represents brain EC in directed graph $\mathcal{G}$ estimated by FSTA-EC, in other words, $\mathbf{A}_{ij}$ symbolizes the effective connectivity from brain region $v_j$ to brain region $v_i$. 
We then apply the softmax function to activate the brain EC matrix $\mathbf{A}$. This function normalizes each row of the brain EC matrix $\mathbf{A}$, guaranteeing that the sum of each row is 1. State differently, for a given brain region $v_i$, the sum of weighted causal effects exerted by all other brain regions on brain region $v_i$ is equal to 1. Subsequently, brain EC matrix $\mathbf{A}$ is multiplied by temporal attention output $Z_T$ to extract spatiotemporal fMRI features:
\begin{equation}
\hat{\mathcal{X}}=\Phi(\mathbf{A} \mathcal{Z}_{\mathbi{T}}),
\label{get_hat_X}
\end{equation}
where $\Phi(\cdot)$ stands for a sequence of operations within the FFN module. This multiplication operation enables a weighted summation of effectively connected temporal features of brain regions, allowing the extraction of spatiotemporal contextual information through brain EC to effectively direct attention towards neighboring brain region nodes that contain more significant information. Finally, 
we obtain two-dimensional fMRI data $\hat{X}\in\mathbb{R}^{N\times T}$ by
\begin{equation}
\hat{X}=\psi(\otimes \hat{\mathcal{X}}),
\end{equation}
where $\psi(\cdot)$ denotes squeeze operation and $\otimes$ signifies two-dimensional convolution with the kernel size of $1\times1$. 

\subsection{Loss Function}
Our formulated loss function encompasses both the prediction loss and sparsity loss components, rendering its calculation as follows:
\begin{equation}
\mathcal{L}=\| X-\hat{X} \|_2^2+\alpha\|\mathbf{A}_{ij}\|_1,
\end{equation}
where $\alpha$ is a hyperparameter. The initial term of the equation implies the predicted loss of FSTA-EC, while the subsequent term represents the sparsity loss of brain EC. 
The algorithm description and pseudocode can be found in Appendix ~\ref{apd_alg}.

\section{Experimental Setting}
In this section, we introduce the experimental setup from four perspectives: data description, evaluation metrics, parameter settings and analysis, and baseline methods.

\subsection{Data Description}\label{Data Description} 
\textbf{Simulated fMRI Dataset.} 
Since there is no standard answer for real-resting-state fMRI dataset, to verify the validity of FSTA-EC algorithm, we first perform experiments on high-noise simulated fMRI dataset. 
For the simulation dataset, we utilize the Sanchez dataset\footnote{https://github.com/cabal-cmu/feedbackdiscovery} \cite{twoStep}, which is an extension of the Smith dataset \cite{smith}. The Smith dataset is generated through dynamic causal models. However, the Sanchez dataset differs in a significant aspect—it is specifically designed to minimize the impact of non-Gaussian influences on the high-noise blood oxygen level dependent signals.
Each dataset in Sanchez dataset has 60 study subjects and 500 data points per subject. In our experiments, we select the first four simulation cases from the Sanchez dataset. Specifically, Sim1, Sim2, and Sim3 all feature a large loop involving all nodes. Sim1 has a 2-cycle, Sim2 has two 2-cycles with shared nodes, and Sim3 has two 2-cycles without shared nodes, with Sim1, Sim2, and Sim3 consisting of 5 nodes and 6, 7, and 7 edges, respectively. In contrast, Sim4 presents a more intricate situation with 10 nodes and 19 edges.
Specific information about the simulated fMRI dataset is shown in Table~\ref{simulation_dataset}.

\begin{table}
\caption{Description of the benchmark simulation dataset.}
\label{simulation_dataset}
\renewcommand\arraystretch{0.8}
\begin{threeparttable}
\tabcolsep 0.1 in
\setlength{\tabcolsep}{0.7mm}{ 
\begin{tabular}{c| c c c c c c}\toprule
Dataset&  Nodes& Arcs& Sessions (s)& TR(s)& Frequency(Hz) & SNR(dB)\\\toprule
Sim1&  5&  6 & 10  & 1.2& 1/200 & 2\\ 
Sim2&  5&  7 & 10  & 1.2& 1/200 & 2\\ 
Sim3&  5&  7 & 10  & 1.2& 1/200 & 2\\ 
Sim4&  10&  19 & 10  & 1.2& 1/200 & 2\\ \bottomrule
\end{tabular}}
\end{threeparttable}
\end{table}

\noindent\textbf{Real-resting-state fMRI Dataset.} 
The real-resting-state fMRI dataset\footnote{https://github.com/shahpreya/MTlnet} consists of 23 samples, each containing 421 data points. We consider the seven regions of interest (ROIs) from the medial temporal lobe as described in \cite{shah2018mapping}, specifically CA1 (Cornu Ammonis 1), CA23DG (Cornu Ammonis 2, 3, and Dentate Gyrus), SUB (Subiculum), ERC (Entorhinal Cortex), BA35 (Brodmann Areas 35), BA36 (Brodmann Areas 36), and PHC (Parahippocampal Cortex). These regions are denoted by the numbers 1 to 7, respectively

\subsection{Evaluation Metrics} 
To assess the effectiveness of FSTA-EC, we used the following five evaluation metrics: Precision, Recall, F1, Accuracy, and Structural Hamming distance (SHD). The detailed descriptions are shown in Appendix ~\ref{apd_metrics}.

\begin{table*}[t]
\centering
\caption{The mean and variance of the eight methods on the four Sanchez simulated datasets. The best and second-best values are \textbf{highlighted} and \underline{underlined}.}
\label{result} 
\renewcommand\arraystretch{0.8}
\begin{threeparttable}
\setlength{\tabcolsep}{1.5mm}{ 
\begin{tabular}{r| r| c c c c c c c |c }
\toprule \multirow{3}{*}{Data}& \multirow{3}{*}{Metrics}  &
\multicolumn{8}{c}{Methods}\\\cmidrule{3-10}
  &   & pwLiNGAM &spDCM& lsGC & ACOCTE& RL-EC &CR-VAE & MetaCAE &  FSTA-EC\\
  &  & \citeyear{pw} & \citeyear{spDCM} & \citeyear{lsGC} & \citeyear{ACOCTE} & \citeyear{EC-DRL} & \citeyear{CR-VAE} & 
  \citeyear{MetaCAE} & (\textbf{Ours}) \\ \midrule
\multirow{5}{*}{Sim1}
   &  Precision $\uparrow$ & 0.63$\pm$0.00 & 0.50$\pm$0.00 & 0.67$\pm$0.00 & \textbf{0.79\boldmath $\pm$0.05}  & 0.66$\pm$0.11 & 0.53$\pm$0.14 & 0.75$\pm$0.09 & \underline{0.78$\pm$0.11} \\
   &  Recall $\uparrow$ & \textbf{0.83\boldmath $\pm$0.00} & 0.33$\pm$0.00 & 0.67$\pm$0.00 & 0.67$\pm$0.07  & 0.35$\pm$0.04 & 0.72$\pm$0.08 & 0.50$\pm$0.01 & \underline{0.78$\pm$0.11} \\
   & F1 $\uparrow$ & 0.71$\pm$0.00 & 0.40$\pm$0.00 & 0.67$\pm$0.00 & \underline{0.73$\pm$0.04}   & 0.46$\pm$0.05 & 0.60$\pm$0.12 & 0.60$\pm$0.08 & \textbf{0.77\boldmath $\pm$0.09} \\
   & Accuracy $\uparrow$ & 0.84$\pm$0.00 & 0.76$\pm$0.00 & 0.84$\pm$0.00 &  \underline{0.88$\pm$0.03}  & 0.81$\pm$0.09& 0.76$\pm$0.10 & 0.84$\pm$0.10 & \textbf{0.89\boldmath $\pm$0.04} \\
   & SHD $\downarrow$ & 4.00$\pm$0.00 & 6.00$\pm$0.00 & 4.00$\pm$0.00 &  \underline{3.00$\pm$0.66} & 5.24$\pm$0.85& 6.24$\pm$2.45 & 4.05$\pm$0.52 & \textbf{2.70\boldmath $\pm$1.00} \\
\midrule
\multirow{5}{*}{Sim2}
    &  Precision $\uparrow$ & 0.30$\pm$0.00 & 0.57$\pm$0.00 & 0.60$\pm$0.00 & 0.67$\pm$0.02 &  \underline{0.80$\pm$0.17} & 0.50$\pm$0.17 & 0.50$\pm$0.11 & \textbf{0.85\boldmath $\pm$0.15} \\ 
    &  Recall $\uparrow$ & \underline{0.86$\pm$0.00} & 0.57$\pm$0.00 & 0.86$\pm$0.00 & 0.57$\pm$0.07 & 0.57$\pm$0.06 & 0.76$\pm$0.05 & 0.29$\pm$0.06 & \textbf{0.87\boldmath $\pm$0.09} \\
    & F1 $\uparrow$ & 0.44$\pm$0.00 & 0.57$\pm$0.00 & \underline{0.71$\pm$0.00} & 0.62$\pm$0.04   & 0.66$\pm$0.08 & 0.60$\pm$0.03 & 0.36$\pm$0.06 & \textbf{0.85\boldmath $\pm$0.11} \\
    & Accuracy $\uparrow$ & 0.40$\pm$0.00 & 0.76$\pm$0.00 & 0.80$\pm$0.00 & 0.80$\pm$0.02 &  \underline{0.84$\pm$0.05} & 0.72$\pm$0.04 & 0.72$\pm$0.10 & \textbf{0.91\boldmath $\pm$0.08} \\
    & SHD $\downarrow$ & 15.00$\pm$0.00 & 6.00$\pm$0.00 & 5.00$\pm$0.00 & 4.96$\pm$0.50 & \underline{4.00$\pm$0.43}  & 7.08$\pm$0.98 & 7.00$\pm$0.26 & \textbf{2.20\boldmath $\pm$1.94} \\
 \midrule
\multirow{5}{*}{Sim3}
   &  Precision $\uparrow$ & 0.67$\pm$0.00 & 0.50$\pm$0.00 & 0.75$\pm$0.00 & 0.81$\pm$0.04  & \textbf{1.00\boldmath $\pm$0.04} & 0.74$\pm$0.04 & 0.33$\pm$0.14 & \underline{0.96$\pm$0.09} \\
   &  Recall $\uparrow$ & \underline{0.86$\pm$0.00} & 0.43$\pm$0.00 & 0.86$\pm$0.00 & 0.57$\pm$0.15  & 0.56$\pm$0.03 & \textbf{0.90\boldmath $\pm$0.14} & 0.14$\pm$0.18 & 0.61$\pm$0.07 \\
   & F1 $\uparrow$ & 0.75$\pm$0.00 & 0.46$\pm$0.00 & \underline{0.80$\pm$0.00} & 0.67$\pm$0.03   & 0.72$\pm$0.03 & \textbf{0.81\boldmath $\pm$0.07} & 0.21$\pm$0.11 & 0.74$\pm$0.06 \\
   & Accuracy $\uparrow$ & 0.84$\pm$0.00 & 0.72$\pm$0.00 & 0.87$\pm$0.00 & 0.84$\pm$0.03  & \underline{0.88$\pm$0.07}& 0.87$\pm$0.03 & 0.68$\pm$0.09 & \textbf{0.88\boldmath $\pm$0.03} \\
   & SHD $\downarrow$ & 4.00$\pm$0.00 & 7.00$\pm$0.00 & \underline{3.00$\pm$0.00} & 4.00$\pm$0.60  & 3.07$\pm$0.46 & 3.00$\pm$0.87 & 8.03$\pm$0.48 & \textbf{2.95\boldmath $\pm$0.74} \\
\midrule
\multirow{5}{*}{Sim4}
    &  Precision $\uparrow$ & 0.32$\pm$0.00 & 0.49$\pm$0.00 & 0.25$\pm$0.00 & 0.54$\pm$0.05 & \textbf{1.00\boldmath $\pm$0.02}  & 0.70$\pm$0.03 & 0.53$\pm$0.12 & \underline{0.75$\pm$0.13} \\
    &  Recall $\uparrow$ & 0.74$\pm$0.00 & \textbf{0.89\boldmath $\pm$0.00} & \underline{0.89$\pm$0.00} & 0.32$\pm$0.03 & 0.42$\pm$0.04  & 0.65$\pm$0.03 & \underline{0.12$\pm$0.03} & 0.59$\pm$0.03 \\
    & F1 $\uparrow$ & 0.44$\pm$0.00 & 0.63$\pm$0.00 & 0.40$\pm$0.00 & 0.39$\pm$0.03   & 0.59$\pm$0.03 & \underline{0.64$\pm$0.03} & 0.20$\pm$0.04 & \textbf{0.66\boldmath $\pm$0.06} \\
    & Accuracy $\uparrow$ & 0.65$\pm$0.00 & 0.80$\pm$0.00 & 0.49$\pm$0.00 & 0.82$\pm$0.01 & 0.88$\pm$0.16  & \underline{0.86$\pm$0.01} & 0.81$\pm$0.07 & \textbf{0.88\boldmath $\pm$0.03} \\
    & SHD $\downarrow$ & 35.00$\pm$0.00 & 20.00$\pm$0.00 & 51.00$\pm$0.00 & 18.37$\pm$1.19 & \underline{12.15$\pm$1.59 }  & 14.05$\pm$1.07 & 18.86$\pm$0.11 & \textbf{11.85\boldmath $\pm$2.76} \\
 \bottomrule
\end{tabular} } 
\end{threeparttable} 
\end{table*}

\subsection{Parameter Setting and Analysis}\label{parameter_setting_section}
After FSTA-EC output brain EC matrix $\mathbf{A}$, a threshold $\theta$ is needed to make $\mathbf{A}$ become a binary matrix. Considering that different parameters are needed for different data, the threshold $\theta$ is set to be adaptive in this paper, and the threshold $\theta$ can be calculated by:
\begin{equation}
\theta=min(|\mathbf{A}|)+ \eta\times(max(|\mathbf{A}|)-min(|\mathbf{A}|)),
\end{equation}
where $\eta$ is a hyperparameter used to adjust thresholds.
Our model implementation is on PyTorch \cite{paszke2019pytorch} with all experiments conducted on Nvidia RTX 3090 (24GB) GPUs. The time complexity of Fourier attention is $O(NlogN)$. This includes the fast Fourier transform (FFT), learnable filter $\mathcal{K}$, inverse fast Fourier transform (IFFT), and feedforward network (FFN) modules. Meanwhile, the time complexity of temporal attention is $O(T^2)$, and the time complexity of spatiotemporal fusion attention is $O(N^2)$. Therefore, the overall time complexity of spatiotemporal attention is $O(T^2+N^2)$. Overall, the time complexity of the Fourier spatiotemporal attention (FSTA-EC) is $O(NlogN+T^2+N^2)$. For the first simulated fMRI dataset, training FSTA-EC takes approximately 44 seconds.

In the training process of FSTA-EC, the training epoch is 300 and the batch size is 32. The seed is set to 42 to ensure reproducibility. We adopt Adam optimizer with the momentum terms $\beta_1$ = 0.90 and $\beta_2$ = 0.98. The embedding size in the input embedding is set to 16 and the dropout is configured to 0.2. The attention in STA with 2 heads.

\subsection{Baseline Methods}
Seven baseline methods are used for comparison with FSTA-EC, which can be categorized into two groups: model-based methods and data-driven methods. The baseline methods as follows: pwLiNGAM \cite{pw}, spDCM \cite{spDCM}, lsGC \cite{lsGC}, ACOCTE \cite{ACOCTE}, RL-EC \cite{EC-DRL}, CR-VAE \cite{CR-VAE}, MetaCAE \cite{MetaCAE}.  The parameter settings for these baseline methods are shown in Table~\ref{parameter_setting}.

\begin{table}[h]
\caption{Parameter settings of seven baseline methods.}
\label{parameter_setting}
\renewcommand\arraystretch{0.5}
\scalebox{0.9}{
\setlength{\tabcolsep}{1.mm}{ 
\begin{tabular}{c| c| c }\toprule
Methods&    Years&Parameters\\ \toprule
pwLiNGAM& 2010 & method = 1 \\
\\
\multirow{2}{*}{spDCM} & \multirow{2}{*}{2014} &${\rm nonlinear}=0,{\rm two\_\,state}=0, {\rm stochastic}=1$, \\
& &{\rm centre}=1, {\rm induced}=1, {\rm maxit}=10\\
\\
lsGC &2017&${\rm cmp}=5,{\rm ARorder}=2,{\rm normalize}=1 $\\
\\
ACOCTE& 2022 &  $\alpha=1.0,\beta=2.0,q_0=0.98,\rho=0.2$\\
\\
RL-EC& 2022 & ${\rm nh}=256,{\rm heads}=16,{\rm stacks}=6$, ${\rm nh}_{\rm decoder}=16$\\
\\
CR-VAE& 2023& ${\rm context}=20, \lambda=0.1, {\rm lr}=0.05$,${\rm nh}=64$ \\
\\
\multirow{2}{*}{MetaCAE} & \multirow{2}{*}{2024} &${\rm nh}=64,\alpha=0.05,\beta=20.0,{\rm k}=3,{\rm d}=4,$\\
& & ${\rm lr}_1=0.02,{\rm lr}_2=0.02,{\rm lr}_3=0.001,{\rm lr}_{\rm main}= 0.002$ \\
\\\bottomrule
\end{tabular}}
}
\end{table}

\section{Experimental Results and Discussions}
In the experiments, we first compare state-of-the-art methods using simulated fMRI data from known ground-truth networks. Then, we demonstrate the practical application of FSTA-EC to publicly available real-resting-state fMRI data, followed by downstream classification tasks on disease datasets. Finally, we conduct an analysis of the model's modules, hyperparameters, and efficiency.

\begin{figure*}[t]
\begin{center}
\includegraphics[width=0.95\textwidth, height=0.12\textheight]{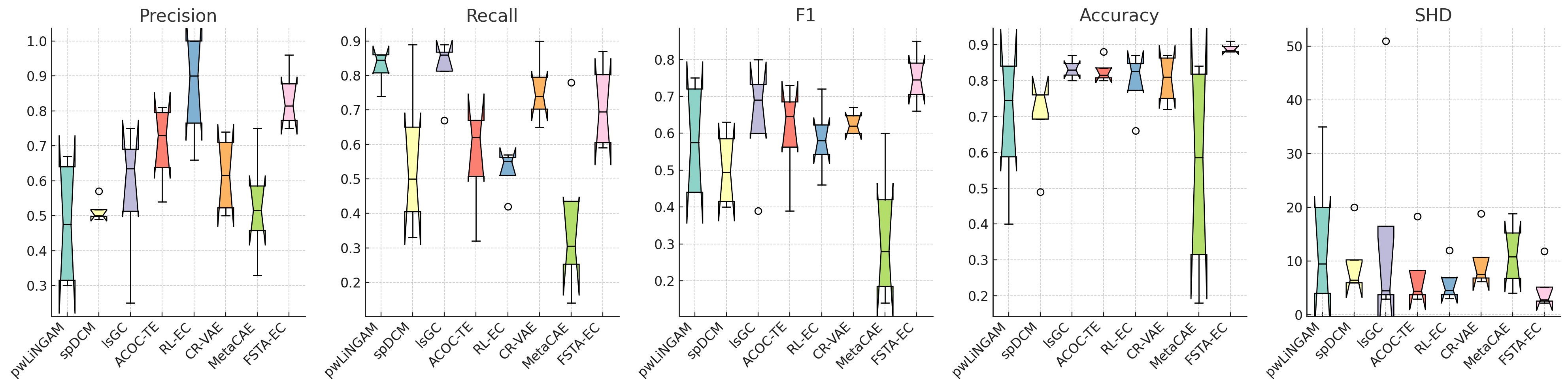}
\end{center}
\caption{Boxplots of the eight methods for each evaluation metric on the four Sanchez simulated datasets.}
\label{sanch_box}
\end{figure*}

\begin{figure*}[t]
	\centering
	\subfigure[pwLiNGAM]{
		\begin{minipage}[ht]{0.23\linewidth}
			\centering
			\includegraphics[width=1.3in]{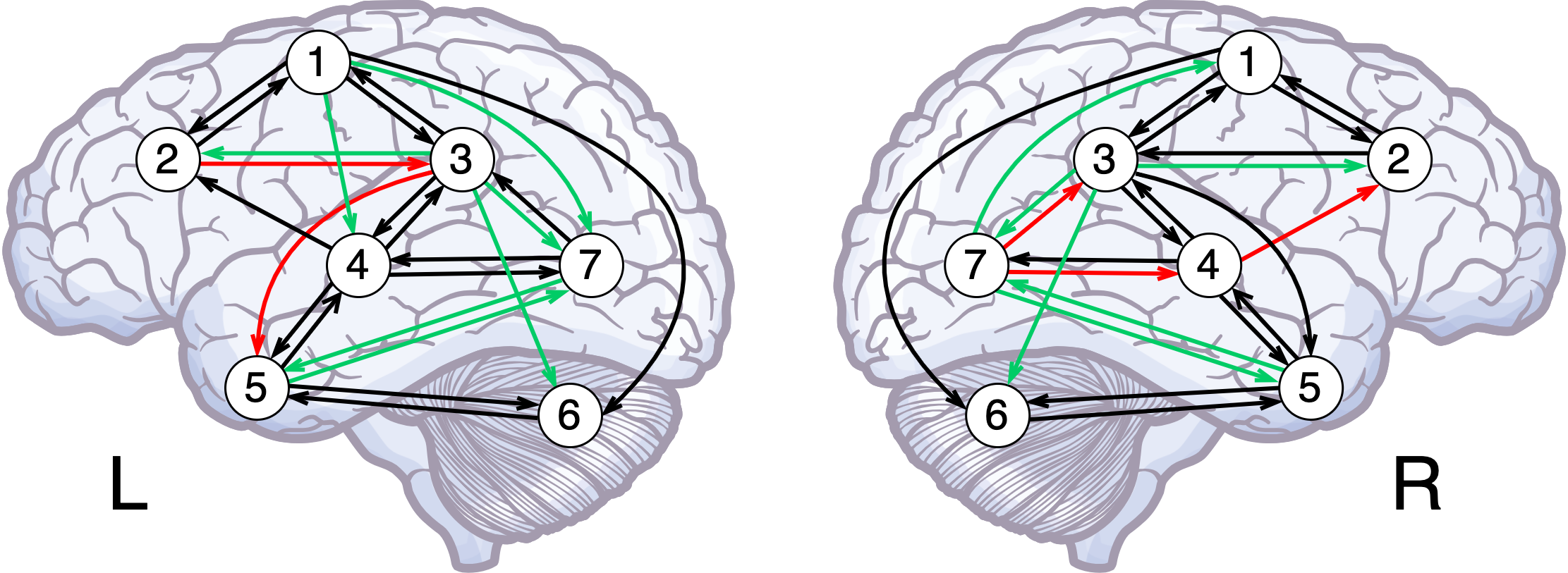}
		\end{minipage}
	}%
	\subfigure[spDCM]{
		\begin{minipage}[ht]{0.23\linewidth}
			\centering
			\includegraphics[width=1.3in]{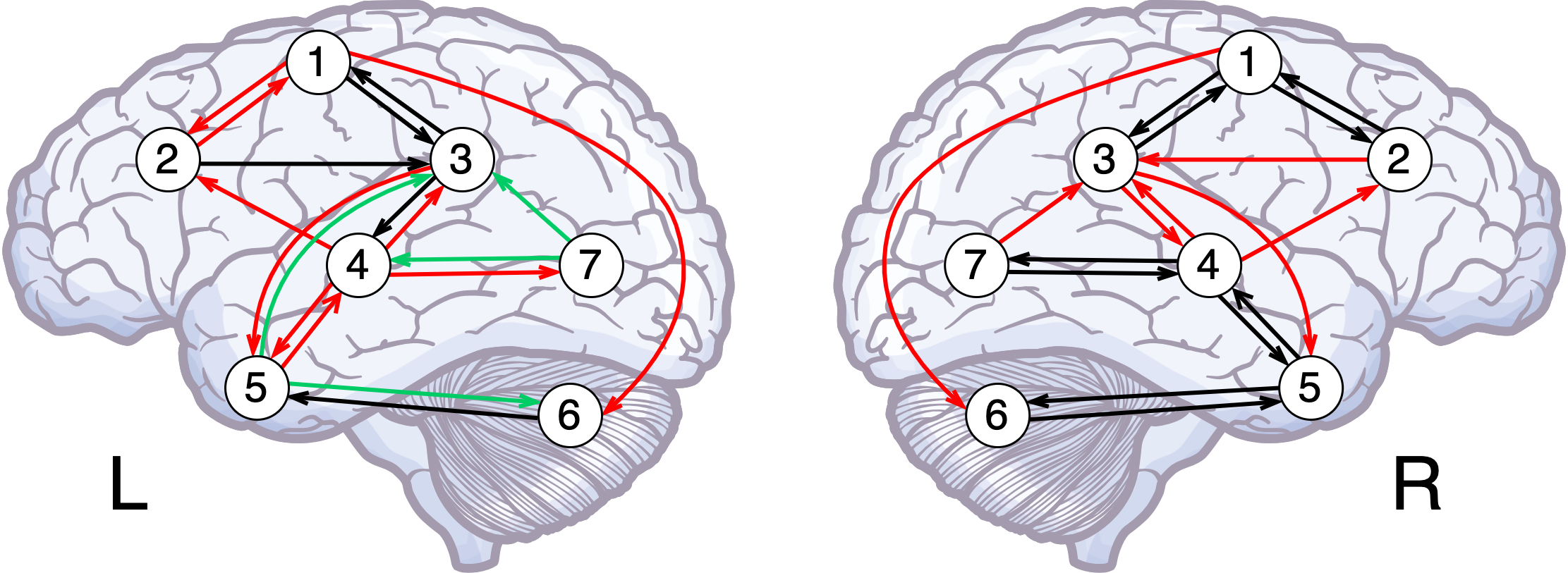}
		\end{minipage}
	}%
        \subfigure[lsGC]{
		\begin{minipage}[ht]{0.23\linewidth}
			\centering
			\includegraphics[width=1.3in]{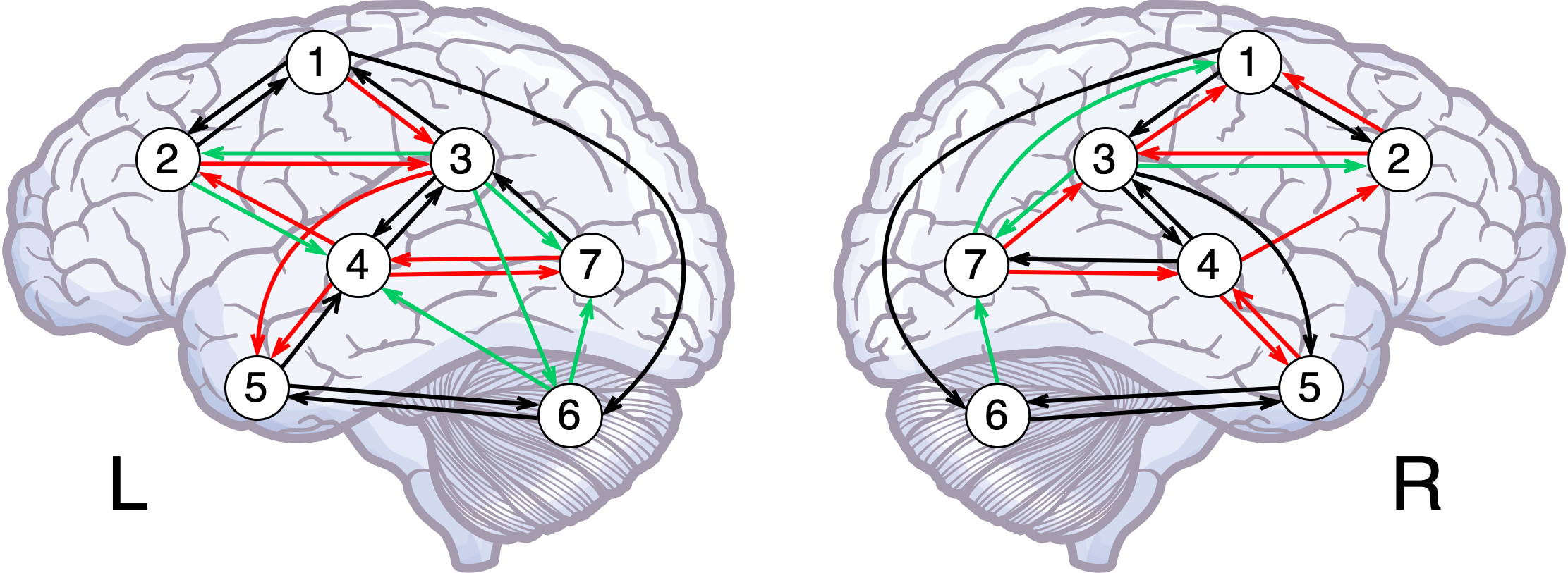}
		\end{minipage}
	}%
	\subfigure[ACOCTE]{
		\begin{minipage}[ht]{0.23\linewidth} 
			\centering
			\includegraphics[width=1.3in]{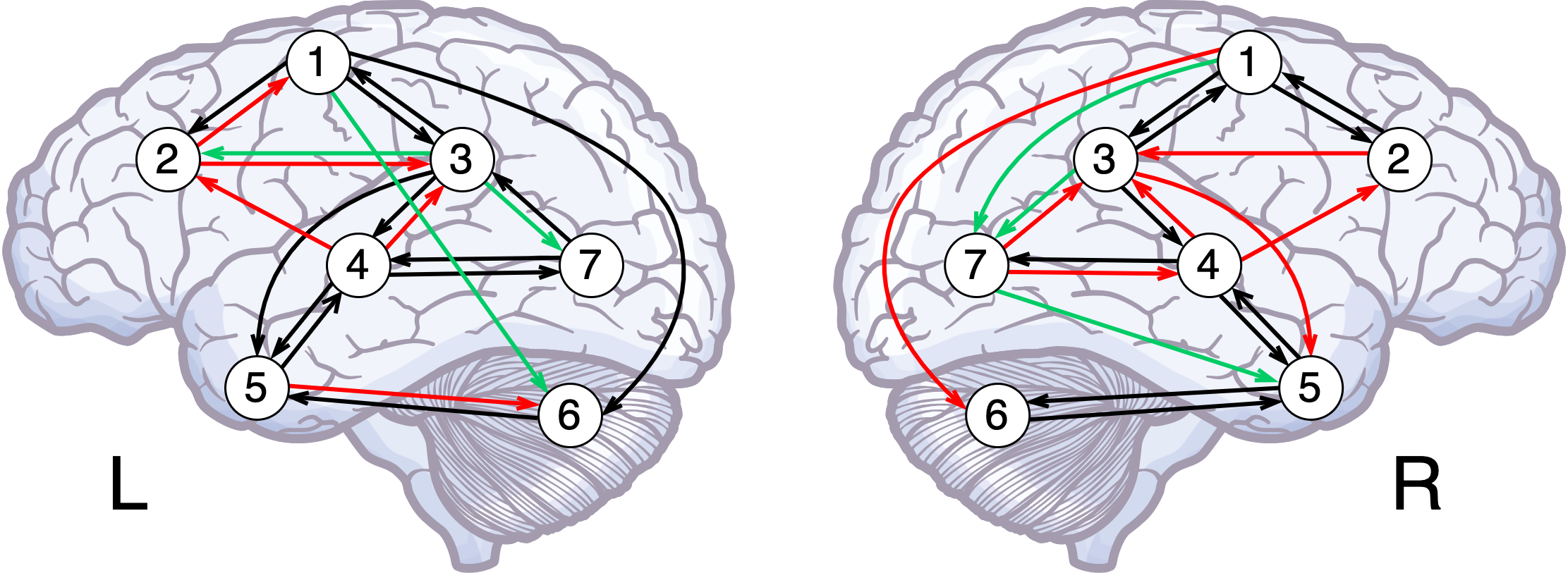} 
		\end{minipage}
	}%
 
	\subfigure[RL-EC]{
		\begin{minipage}[ht]{0.23\linewidth}
			\centering
			\includegraphics[width=1.3in]{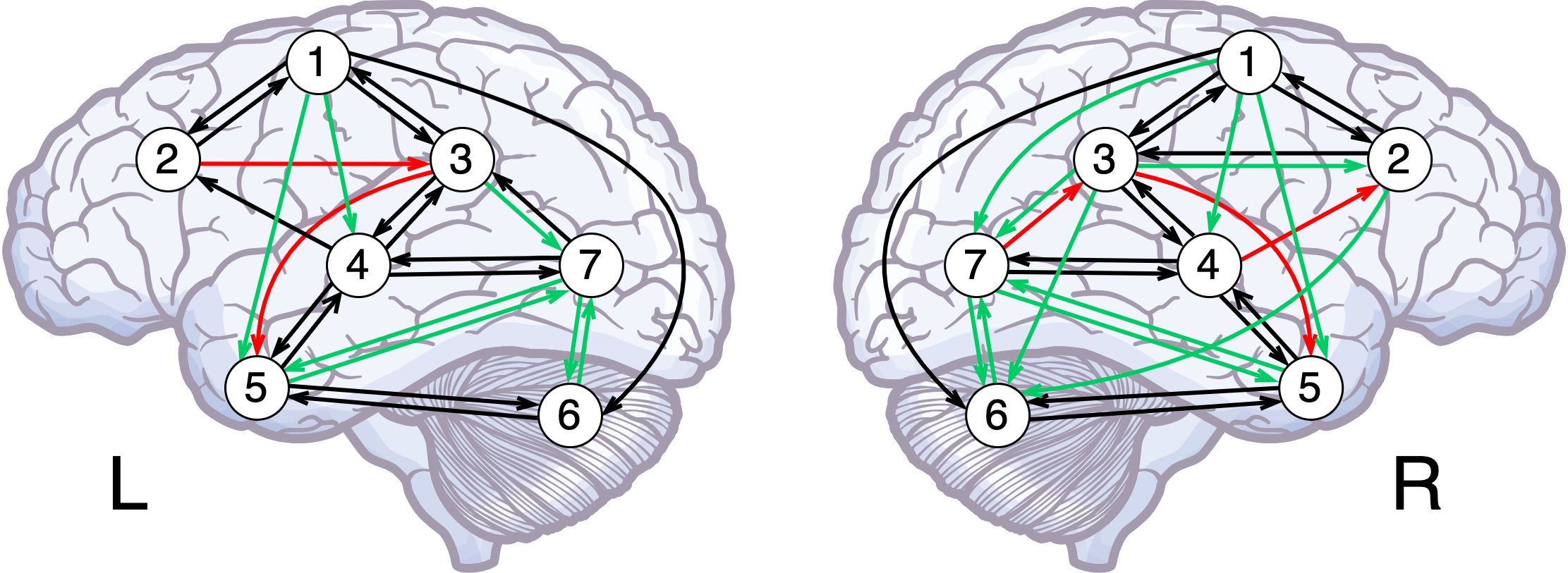}
		\end{minipage}
	}%
	\subfigure[CR-VAE]{
		\begin{minipage}[ht]{0.23\linewidth}
			\centering
			\includegraphics[width=1.3in]{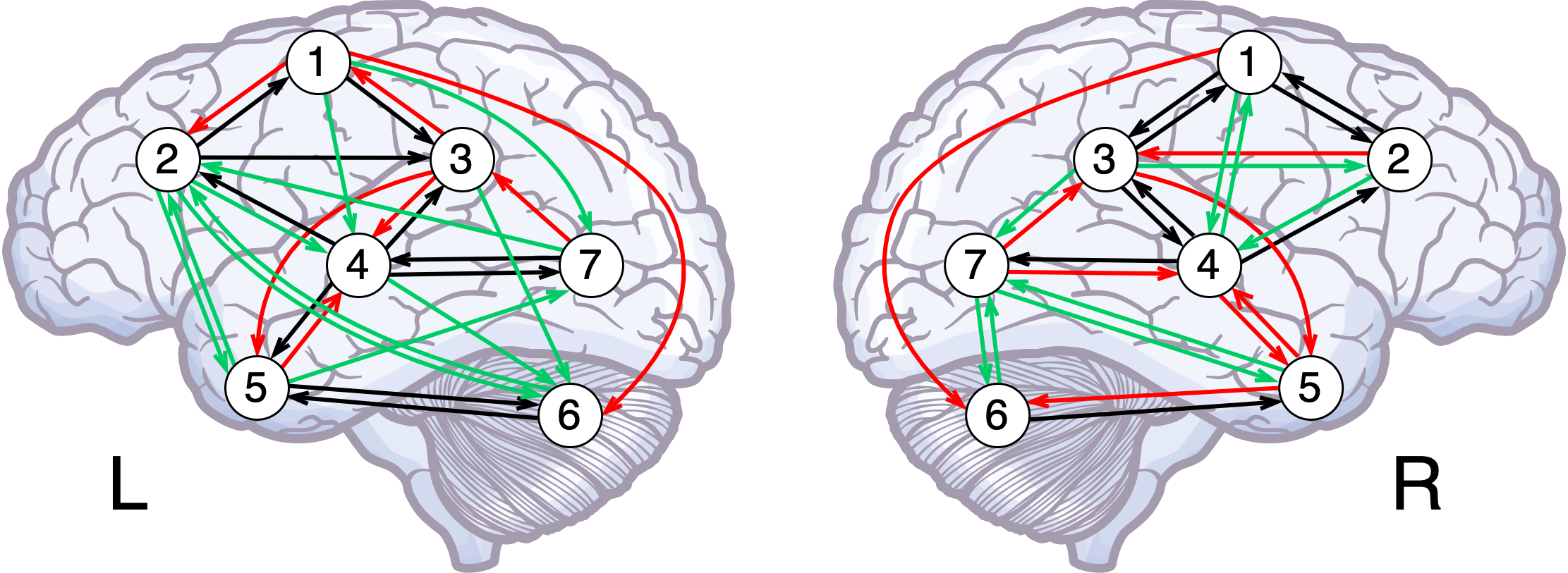}
		\end{minipage}
	}%
        \subfigure[MetaCAE]{
		\begin{minipage}[ht]{0.23\linewidth}
			\centering
			\includegraphics[width=1.3in]{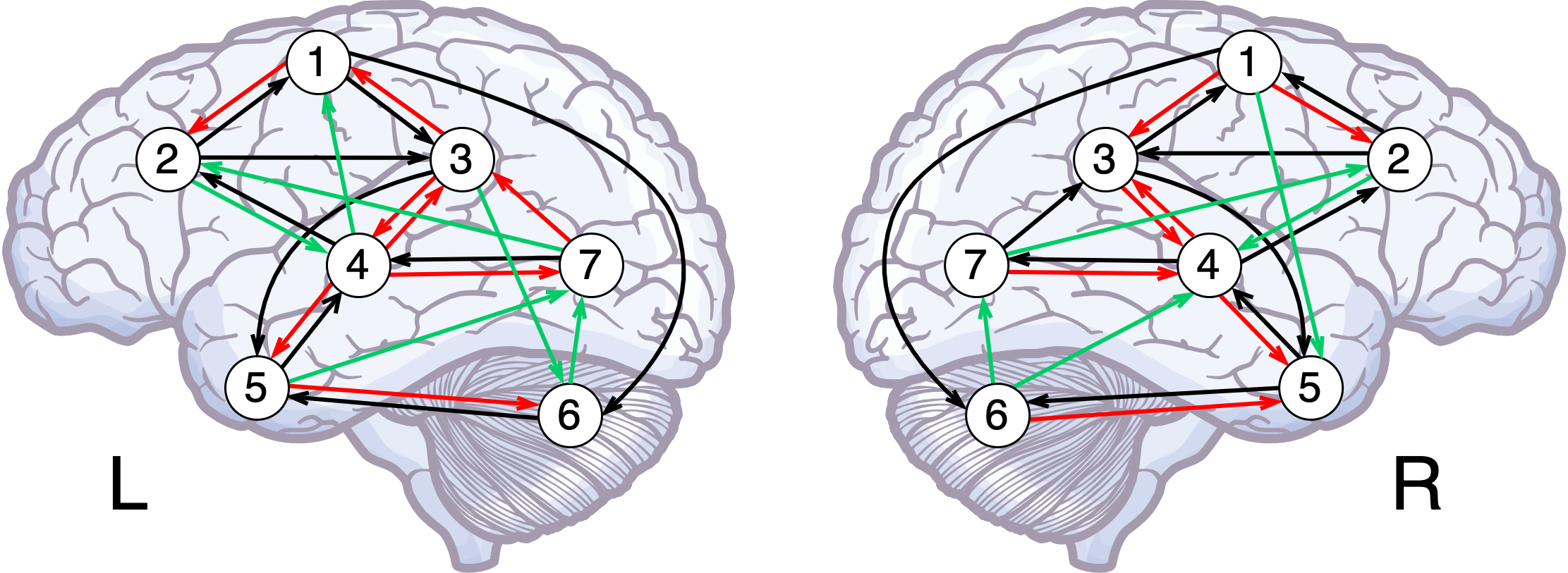}
		\end{minipage}
	}%
	\subfigure[FSTA-EC]{
		\begin{minipage}[ht]{0.23\linewidth}
			\centering
			\includegraphics[width=1.3in]{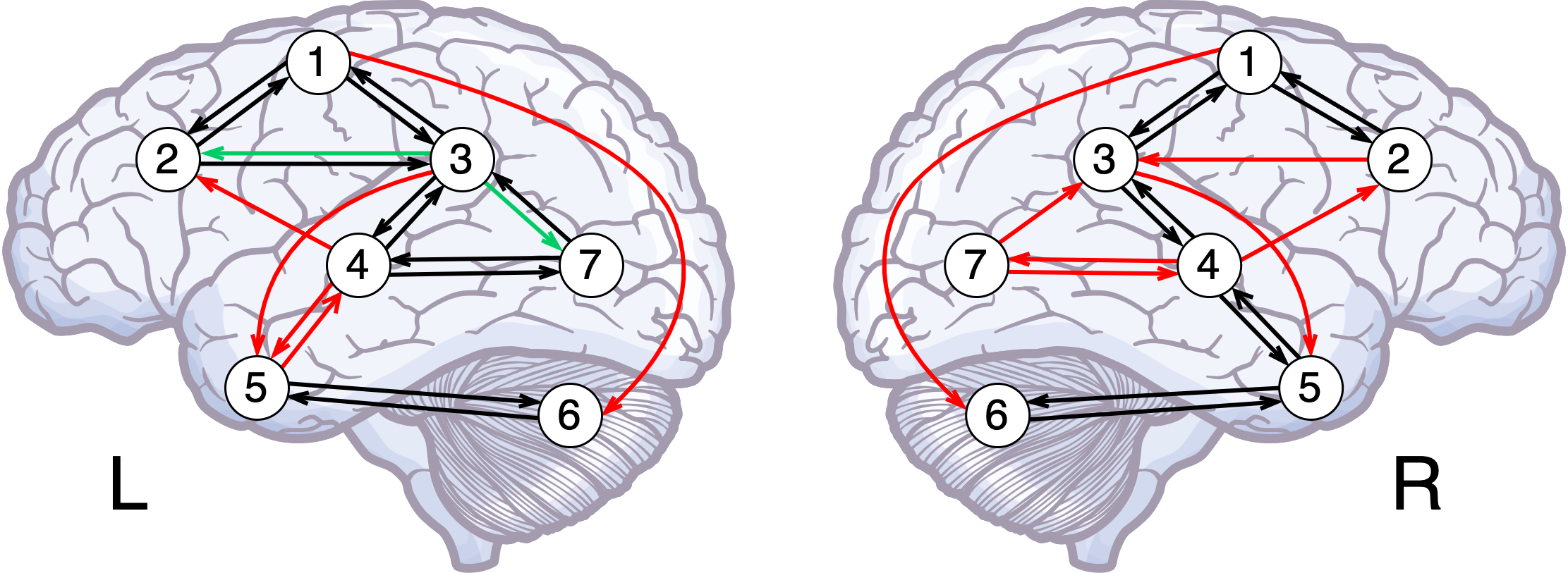}
		\end{minipage}
	}%
	\centering
	\caption{Brain effective connectivity estimated by seven baseline methods and FSTA-EC.}
	\label{EC}
\end{figure*}

\subsection{Results on Simulated fMRI Dataset}\label{Results on Simulated fMRI}
In experiments on simulated fMRI datasets with Gaussian noise,
we run FSTA-EC and 7 baseline methods 20 times on all subjects in each simulation (group analysis) and calculate the mean and variance over 20 implementations. The obtained results are presented in Table ~\ref{result}. Note that some methods yield the same results in multiple runs, thus the variance is 0. For additional comparative experiments, please refer to Appendix ~\ref{apd_additional_exper}.

From the results of Sim1, with the exception of pwLiNGAM, ACOCTE, and FSTA-EC, all other methods performed poorly, with FSTA-EC demonstrating the best overall performance. 

Sim2 and Sim3 have an additional 2-cycle compared to Sim1. Specifically, the two 2-cycles of Sim2 share a node, while the two 2-cycles of Sim3 do not share a node. In Sim2, FSTA-EC outperforms all other methods across every evaluation metric. 
In Sim3, CR-VAE achieves the best results in Recall and F1. RL-EC achieves high Precision; however, its Recall performance is relatively low. In terms of MetaCAE's performance, it shows lower scores across all metrics in Sim2 and Sim3. On the other hand, FSTA-EC outshines as it achieves the highest performance across multiple evaluation metrics.

Compared to Sim1, Sim2, and Sim3, Sim4 has 10 nodes. The majority of methods exhibit excellent performance in the first three simulations, but their performance is comparatively weaker in Sim4, possibly due to the increased complexity of the network with 10 nodes compared to the previous three simulations, which only featured 5 nodes. Furthermore, MetaCAE as a small-sample method, does not demonstrate an advantage in group analysis. In particular, FSTA-EC still outperforms other baseline methods. 


To facilitate a more comprehensive comparison of these methods' performance, we calculate the evaluation metrics of these methods using four simulated fMRI datasets above and plot the boxplot, which is depicted in Figure ~\ref{sanch_box}. 
It can be seen that FSTA-EC outperforms the other methods in F1, Accuracy, and SHD. While RL-EC achieves the highest Precision, this is likely due to its high Precision but low Recall in Sim3 and Sim4. Although pwLiNGAM and lsGC perform well in Recall, they fall short in other evaluation metrics. Overall, FSTA-EC stands out as the best performing method, surpassing the benchmark methods and demonstrating superior performance.

\begin{table}[htbp] \centering
\caption{$p$-values obtained from the t-test for FSTA-EC and seven baseline methods. Underlined values indicate no significant difference at the 95\% confidence level.}
\label{t_test}
\begin{threeparttable}
\resizebox{1.0\linewidth}{!}{
\begin{tabular}{c |c c c c c}\toprule
Methods  & Precision & Recall & F1 & Accuracy & SHD\\ \toprule
pwLiNGAM     & $1.54\times 10^{-30}$ & $1.10\times 10^{-9}$  & $9.07\times 10^{-13}$ & $1.59\times 10^{-23}$ & $1.73\times 10^{-7}$ \\
spDCM     & $2.08\times 10^{-16}$ & $2.26\times 10^{-22}$  & $9.34\times 10^{-22}$ & $2.87\times 10^{-21}$ & $9.03\times 10^{-6}$ \\
lsGC  & $1.18\times 10^{-18}$ & $6.16\times 10^{-8}$  & $1.22\times 10^{-7}$ & $1.88\times 10^{-12}$ & $7.96\times 10^{-6}$ \\
ACOCTE & $9.10\times 10^{-11}$ & $2.76\times 10^{-3}$  & $7.29\times 10^{-6}$ & $5.18\times 10^{-4}$ & $8.41\times 10^{-8}$ \\
RL-EC & $\underline{1.13\times 10^{-1}}$ & $8.20\times 10^{-5}$  & $2.92\times 10^{-7}$ & $\underline{5.99\times 10^{-2}}$ & $4.33\times 10^{-2}$ \\
CR-VAE & $5.58\times 10^{-18}$ & $\underline{7.84\times 10^{-1}}$  & $\underline{6.96\times 10^{-2}}$ & $1.23\times 10^{-4}$ & $3.20\times 10^{-9}$ \\
MetaCAE &$9.25\times 10^{-22}$ & $3.56\times 10^{-34}$& $1.87\times 10^{-28}$ & $4.38\times 10^{-15}$ & $8.54\times 10^{-6}$ \\\bottomrule
\end{tabular}}
\end{threeparttable}  
\end{table}

\textbf{T-test results for FSTA-EC and eight baseline methods.}
To delve deeper into the notable distinctions between FSTA-EC and other baseline methods, we conduct a t-test comparing FSTA-EC against the remaining baseline methods. The $p$-value lower than 0.05 implies a significant difference at a 95\% confidence level. The results are shown in Table ~\ref{t_test}. The results reveal significant disparities between FSTA-EC and the majority of the methods across all five evaluation metrics. 

\subsection{Results on Real-resting-state fMRI Dataset}
Since \textbf{none} of the existing real-resting-state fMRI data has ground truths for brain EC network, we can only use the existing work \cite{twoStep} as a reference for evaluating FSTA-EC. 
We conduct experiments utilizing FSTA-EC along with other baseline methods for the left and right hemisphere groups. The resulting EC can be observed in Figure ~\ref{EC}. 
In Figure ~\ref{EC}, the \textbf{black} lines represent correct connections, the \textbf{green} lines stand for spurious connections, and the \textbf{red} lines indicate missing connections.

\begin{figure}[htbp]
\centering
\includegraphics[width=0.5\textwidth]{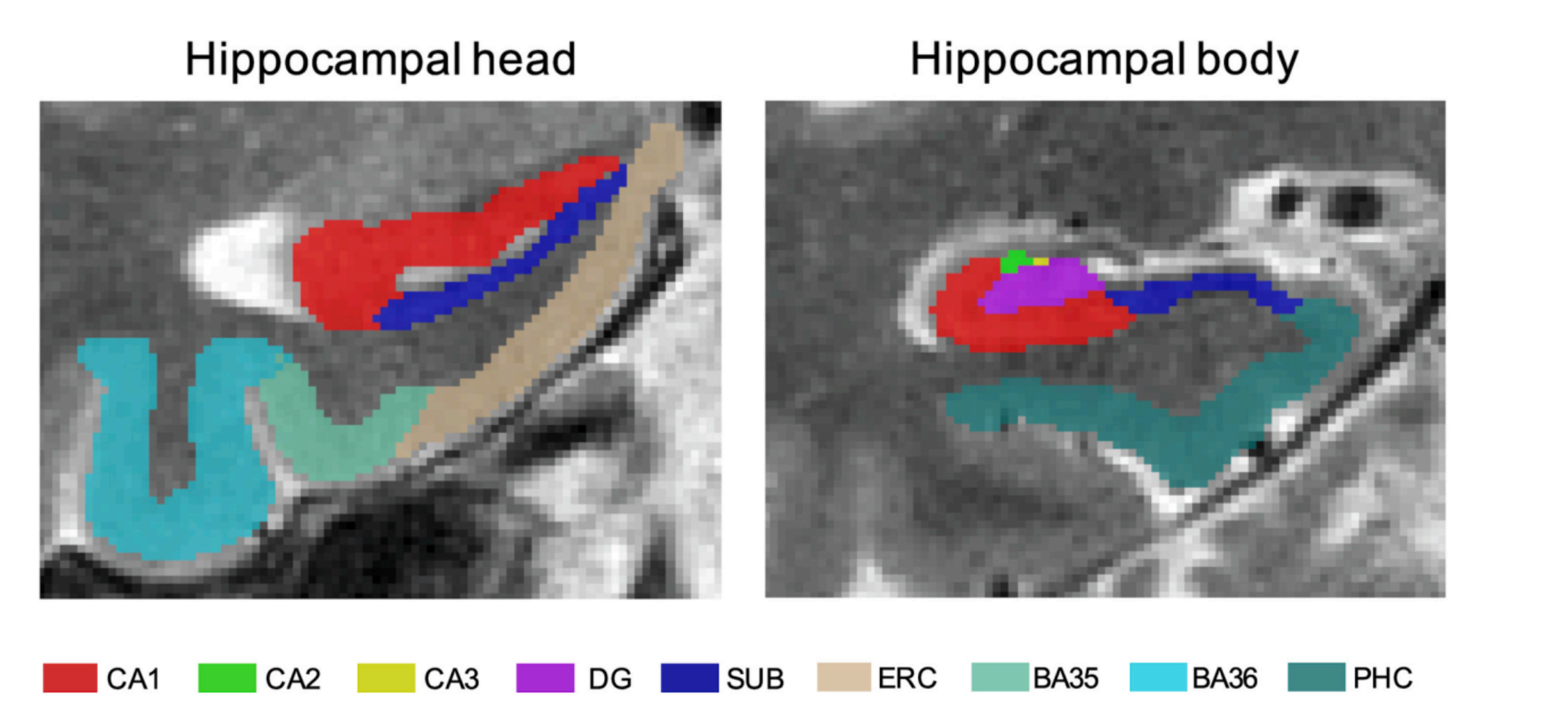}
\caption{Visualization of ROIs for real fMRI dataset in the hippocampal head and hippocampal body  \protect\cite{de2023respective}. Abbreviation: CA1 (Cornu Ammonis 1), CA2 (Cornu Ammonis 2), CA3 (Cornu Ammonis 3), DG (Dentate Gyrus), SUB (Subiculum), ERC (Entorhinal Cortex), BA35 (Brodmann Areas 35), BA36 (Brodmann Areas 36), and PHC (Parahippocampal Cortex). }
\label{ROI}
\end{figure}

From Figure ~\ref{EC}, it is apparent that in the left hemisphere of the brain region, FSTA-EC successfully identifies 12 correct connections, generates 2 spurious connections, and fails to detect 5 correct connections. Similarly, in the right hemisphere, FSTA-EC identifies 10 correct connections, produces no spurious connections, but misses 7 correct connections. Compared to FSTA-EC, other methods exhibit a significantly higher number of false connections and undetected correct connections.
In brief, when considering both the left and right hemispheres, FSTA-EC excels in both identifying the highest number of correct connections and minimizing the generation of spurious connections, setting it apart from other methods.


Figure ~\ref{ROI} presents a visualization of the regions of interest (ROIs) in the real fMRI dataset~\cite{de2023respective}, highlighting their close association with the hippocampus. These ROIs play a crucial role in establishing connections with the hippocampus, facilitating the transmission of information, and contributing to various related functions. In the subsequent analysis, we focus on selecting key effective connectivity to evaluate the performance of each method. Note that the correspondence between the brain EC shown in Figure ~\ref{EC} and the ROIs is explained in the description of the real-resting-state fMRI dataset provided in Section ~\ref{Data Description}.

\noindent\textbf{BA35 $\leftrightarrow$ BA36. }
The Perirhinal Cortex (PRC), situated within anterior medial temporal lobe (MTL), encompasses Brodmann Areas 35 and 36 (BA35 and BA36) \cite{ding2010borders}. The PRC is a crucial cortical region involved in various cognitive functions, such as episodic memory, semantic memory, and visual perceptual processing systems \cite{xie2017multi}. Hence, this two-way pathway is of significant importance. As depicted in the figure, it is obvious that the majority of the methods successfully and accurately estimate this bidirectional pathway.

\noindent\textbf{BA35 $\leftrightarrow$ ERC $\leftrightarrow$ SUB $\leftrightarrow$ CA1. }Apart from BA35 $\leftrightarrow$ ERC connection in left hemisphere medial temporal lobe, FSTA-EC accurately estimates BA35 $\leftrightarrow$ ERC $\leftrightarrow$ SUB $\leftrightarrow$ CA1 in both the left and right hemispheres, aligning with the information flow from PRC to the Entorhinal Cortex (ERC), the Subiculum (SUB) and Cornu Ammonis 1 (CA1) \cite{van1982parahippocampal}. Furthermore, pwLiNGAM, ACOCTE, and RL-EC also demonstrate reasonable accuracy in estimating the effective connectivity of this information flow. However, MetaCAE could only correctly estimate the one-way effective connectivity, possibly due to the penalty imposed by the directed acyclic graph, which restricts the model from identifying bidirectional effective connectivities.

\noindent\textbf{CA1 $\leftrightarrow$ CA23DG. }
The connectivity between CA1 and CA23DG plays a crucial role in supporting episodic memory and provides direct insights into the impact of hydrocortisone on the hippocampus \cite{sherman2023hippocampal}. FSTA-EC, pwLiNGAM, and RL-EC all accurately estimate the effective connectivity. Similarly, MetaCAE continues to exhibit limitations by only estimating unidirectional effective connectivity.

\noindent\textbf{ERC $\rightarrow$ CA23DG. }
This pathway serves as the primary connection between the medial temporal lobe cortices and the hippocampus. However, FSTA-EC fails to capture the connection in the medial temporal lobes of both the left and right hemispheres. Only CR-VAE and MetaCAE successfully capture it on both hemispheres, while pwLiNGAM and RL-EC capture this connection on one hemisphere. Other methods, however, do not capture this specific effective connectivity. A potential explanation for this discrepancy could be caused by the sparsity penalty hyperparameter set too high in the loss function of FSTA-EC. Consequently, the EC obtained by FSTA-EC becomes a sparse matrix, leading to the omission of several connections and reducing the occurrence of spurious connections.

\subsection{Downstream Tasks (Brain Disease Classification using EC networks)}
To further validate the effectiveness of FSTA-EC, we utilize brain EC networks derived from the ADNI\footnote{https://adni.loni.usc.edu/} and ABIDE I\footnote{http://preprocessed-connectomes-project.org/abide/} datasets for downstream classification tasks, comparing the performance of FSTA-EC against other state-of-the-art methods.
In the classification experiment, we can employ the label (AD or HC, ASD or HC) as the groundtruth. If the brain network effective connectivity network estimated by a method can be successfully classified, then we think that this method performs well on effective connectivity estimation. In detail, we compare our proposed method FSTA-EC with several other effective connectivity estimation methods (pwLiNGAM, spDCM, lsGC, ACOCTE, RL-EC, CR-VAE, MetaCAE) for AD classification on the ADNI dataset, and all methods employ the same classifier (random forest with 1000 decision trees). We divided the dataset into 10 parts, nine of them were taken as training data in turn, and one was used as test data for testing. Then all methods run on the same training data and test data.

To analyze the classification performance more comprehensively, we also employ the precision, recall, F1 value to evaluate the performance of different brain effective connectivity learning methods. Table ~\ref{ADNI} and Table ~\ref{ABIDE I} show the mean value and the standard deviation of the 10 runs for each method on ADNI dataset and ABIDE I dataset.

\begin{table}[ht] \centering
\caption{Classification Results on ADNI dataset.}
\label{ADNI}
\begin{threeparttable}
\resizebox{1.0\linewidth}{!}{
\begin{tabular}{c |c c c}\toprule
Methods  & Precision$\uparrow$ & Recall$\uparrow$ & F1$\uparrow$ \\ \toprule
pwLiNGAM     & 73.39$\pm$8.54 & 73.33$\pm$3.33  & 73.35$\pm$4.79  \\
spDCM     & 74.82$\pm$8.60 & 72.67$\pm$5.31  & 73.72$\pm$6.56 \\
lsGC  & \textbf{77.68\boldmath $\pm$5.62} & \underline{74.02$\pm$5.24}  & \underline{75.80$\pm$5.42} \\
ACOCTE & 70.82$\pm$4.17 & 68.96$\pm$2.91  & 69.87$\pm$3.42 \\
RL-EC & 68.60$\pm$4.88 & 67.15$\pm$3.43  & 67.86$\pm$4.02 \\
CR-VAE & 73.12$\pm$7.71 & 70.67$\pm$4.42  & 70.87$\pm$5.61 \\
MetaCAE &71.32$\pm$7.92 & 70.76$\pm$4.14	& 71.03$\pm$5.43 \\
FSTA-EC & \underline{77.21$\pm$6.28} & \textbf{74.67\boldmath $\pm$4.21} & \textbf{75.91\boldmath $\pm$5.04} \\ \bottomrule
\end{tabular}}
\end{threeparttable} 
\end{table}

\begin{table*}[ht]\centering
\caption{Efficiency analysis on Sim1 of Sanchez datasets.}
\label{EfficiencyAnalysis}
\begin{threeparttable}
\begin{tabular}{c |c c c c c c c c c}\toprule
Methods  & pwLiNGAM & spDCM & lsGC & ACOCTE & RL-EC & CR-VAE & MetaCAE & FSTA-EC \\ \toprule
Mem.(MiB/subject) & 19.08 & 13.54 & 13.73 & 136.94 & 32.96 & 50.13 & 199.96 & 25.78 \\
Speed(s/epoch) & 0.27 & 379.58 & 1.04 & 1.97 & 2.85 & 0.02 & 4.48 & 0.15 \\ \bottomrule
\end{tabular}
\end{threeparttable} 
\end{table*}

\begin{table}[ht] \centering
\caption{Classification Results on ABIDE I dataset.}
\label{ABIDE I}
\begin{threeparttable}
\resizebox{1.0\linewidth}{!}{
\begin{tabular}{c |c c c}\toprule
Methods  & Precision$\uparrow$ & Recall$\uparrow$ & F1$\uparrow$ \\ \toprule
pwLiNGAM     & 65.29$\pm$9.97 & \underline{64.13$\pm$7.01}  & \underline{64.70$\pm$7.99}  \\
spDCM     & 64.15$\pm$9.89 & 62.88$\pm$5.10  & 63.51$\pm$7.73 \\
lsGC  & \underline{65.32$\pm$8.92} & 63.01$\pm$4.81  & 64.14$\pm$6.80 \\
ACOCTE & 55.34$\pm$9.51 & 55.67$\pm$3.66  & 55.50$\pm$6.41 \\
RL-EC & 55.76$\pm$9.98 & 53.74$\pm$5.69  & 54.73$\pm$7.22 \\
CR-VAE & 63.12$\pm$9.03 & 63.33$\pm$3.36  & 63.22$\pm$5.89 \\
MetaCAE &62.34$\pm$9.51 & 61.67$\pm$4.31	& 62.00$\pm$5.93 \\
FSTA-EC & \textbf{65.89\boldmath $\pm$6.41} & \textbf{65.57\boldmath $\pm$3.22} & \textbf{65.72\boldmath $\pm$4.65} \\ \bottomrule
\end{tabular}}
\end{threeparttable} 
\end{table}

\subsection{Model Analysis}

\noindent\textbf{Ablation Study}
We perform ablation studies on Fourier attention (FA), temporal attention (TA), the first residual network (Add) 
in FA and the first Add\&Norm in FA, with the results presented in Table ~\ref{ablation}.
As can be seen from Table ~\ref{ablation}, compared with the absence of TA, FA has a significant improvement in all metrics. While TA shows a slight decrease in Recall and F1, the other metrics exhibit enhancements. This further highlights the importance of data denoising and extraction of spatiotemporal correlations. In addition, the Add\&Norm mechanism in FA contributes to a noticeable enhancement in the model's overall performance.

\begin{figure}[htbp]
\centering
\subfigure[Influence of soft threshold hyperparameter $\eta$.]{
\includegraphics[width=8.5cm]{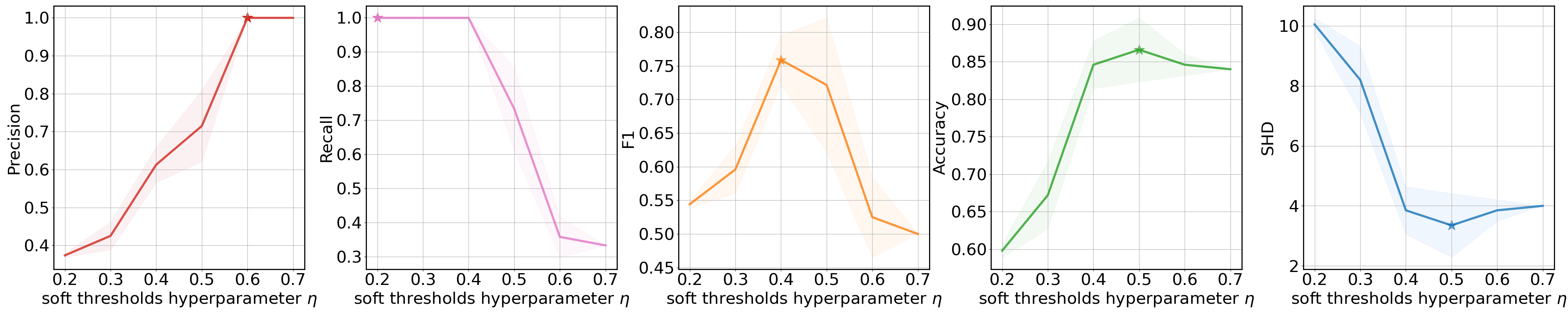}
}

\subfigure[Influence of the number of heads in STA.]{
\includegraphics[width=8.5cm]{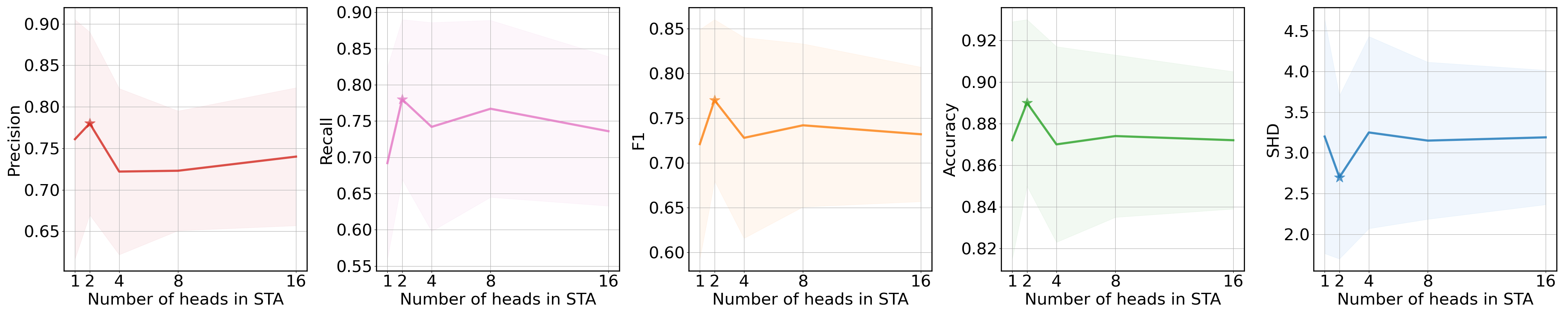}
}
\caption{Hyperparameter analysis on the first simulated fMRI dataset, where the starred results are the best results.}
\label{figure_Hyperparameter_Analysis}
\end{figure}

\noindent\textbf{Hyperparameter Analysis. }
We conduct supplementary experiments on the first simulated fMRI dataset to scrutinize the impact of soft threshold and the number of heads with multiple attention in STA, as illustrated in Figure ~\ref{figure_Hyperparameter_Analysis}. For the chosen evaluation metrics, larger values for the first four metrics indicate better performance, while smaller values for the last metric signify better performance. As depicted in Figure ~\ref{figure_Hyperparameter_Analysis}(a), it is evident that Precision is higher and Recall is lower when the soft threshold hyperparameter $\eta$ exceeds 0.5; conversely, when $\eta$ is below 0.5, Precision decreases while Recall increases. The F1 score offers a comprehensive balance between Precision and Recall, showing that both metrics are higher at a soft threshold hyperparameter $\eta$ of 0.5. Moreover, For Accuracy and SHD, the best performance is achieved when the $\eta$ is 0.5. Hence, after comprehensive consideration, 0.5 is chosen as the final value for the experimental hyperparameter $\eta$. Turning to Figure ~\ref{figure_Hyperparameter_Analysis}(b), it is notable that within STA, selecting 2 heads for multi-head attention results in the highest values for all evaluation metrics.

\noindent\textbf{Efficiency analysis. } 
We compare the running time and memory usage of the baseline methods and FSTA-EC on the first simulated fMRI dataset, and the results are presented in Table ~\ref{EfficiencyAnalysis}. Note that pwLiNGAM, spDCM, and lsGC are MATLAB codes, while ACOCTE is packaged as a jar file. It is evident that FSTA-EC requires significantly less memory compared to ACOCTE and the deep learning algorithms in the baseline method (RL-EC, CR-VAE, MetaCAE). Additionally, its running speed performance is second only to CR-VAE, which may be attributed to the fact that Fourier attention and spatiotemporal attention only have one layer.

\begin{table}[th] 
\centering
\caption{\mbox{Ablation analysis on the Sim1 of Sanchez datasets.}}
\label{ablation}
\renewcommand\arraystretch{1}
\scalebox{0.85}{
\begin{threeparttable}
\setlength{\tabcolsep}{0.8mm}{
\begin{tabular}{c c c c c c c c}
\toprule
\multirow{2}{*}{Variant} & \multicolumn{5}{c}{Metrics}\\
\cmidrule(l{2pt}r{2pt}){2-6}
 & Precision$\uparrow$ & Recall$\uparrow$ & F1$\uparrow$ & Accuracy$\uparrow$ & SHD$\downarrow$ \\
\midrule
\textbf{Default} & \underline{0.78$\pm$0.11} & \underline{0.78$\pm$0.11} & \underline{0.77$\pm$0.09} & \textbf{0.89\boldmath $\pm$0.04} & \textbf{2.70\boldmath $\pm$1.00} \\
w/o FA & \textbf{0.80\boldmath $\pm$0.16} & 0.63$\pm$0.14 & 0.69$\pm$0.11 & 0.87$\pm$0.05 & 3.35$\pm$1.31 \\
w/o TA & 0.74$\pm$0.12 & \textbf{0.85\boldmath $\pm$0.12} & \textbf{0.78\boldmath $\pm$0.10} & \underline{0.88$\pm$0.05} & \underline{2.75$\pm$1.34} \\
w/o FA\&TA & 0.63$\pm$0.13 & 0.67$\pm$0.07 & 0.64$\pm$0.09 & 0.82$\pm$0.05 & 4.50$\pm$1.36 \\
w/o Add & 0.77$\pm$0.16 & 0.73$\pm$0.14 & 0.75$\pm$0.12 & 0.88$\pm$0.05 & 2.80$\pm$1.28 \\
w/o Add\&Norm & 0.74$\pm$0.17 & 0.78$\pm$0.12 & 0.76$\pm$0.12 & 0.87$\pm$0.06 & 3.10$\pm$1.14 \\
\bottomrule
\end{tabular}
}
\end{threeparttable}
}
\end{table}

\section{Conclusion and Future Work}
This paper introduces a novel method, Fourier spatiotemporal attention, for estimating brain effective connectivity (FSTA-EC). FSTA-EC first uses a fast Fourier transform to convert noisy fMRI data into frequency domain, extracting global frequency domain features. Then, it employs temporal attention to learn intricate temporal patterns from fMRI time series data and uses spatiotemporal fusion attention to capture spatiotemporal features and spatial brain effective connectivity among brain regions. Systematic experiments validate the effectiveness and practical significance of the FSTA-EC.

The limitation of FSTA-EC is that the inclusion of dot-product attention in the model imposes certain constraints on its computing speed and memory utilization. As a result, there is room for future optimization of the dot product operation, which will further enhance its overall performance. 

\section*{Acknowledgments}
The authors thank anonymous reviewers for their valuable comments and helpful suggestions. This work was supported by National Natural Science Foundation of China (62106009, 62276010), sponsored by Beijing Nova Program (20240484635), and R$\&$D Program of Beijing Municipal Education Commission (KM202210005030, KZ202210005009).

\clearpage
\newpage

\bibliographystyle{ACM-Reference-Format}
\balance
\bibliography{sample-base}
\appendix

\section{APPENDIX}
\subsection{Notations}\label{apd_notation}
In this paper, we utilize the uppercase font (e.g., $X$) to represent matrix, and the uppercase calligraphic letters (e.g., $\mathcal{X}$) are used to denote three-dimensional (3D) tensor. 
In order to facilitate readers' comprehension of this article, we provide a summary of the symbols used throughout the paper, which is presented in Table ~\ref{notation}.

\begin{table}[bp] \centering
\renewcommand\arraystretch{0.4}
\begin{threeparttable}
\tabcolsep 
0.08 in \footnotesize
\begin{tabular}{c c}\toprule
Symbols&    Meaning\\ \toprule
$X$ & original fMRI data matrix \\
\\
$\hat{X}$ & predicted fMRI data matrix \\
\\
\multirow{2}{*}{$W_{\mathbi{T}}^Q$} & linear transformation weight matrix of Query \\
&in temporal attention\\
\\
\multirow{2}{*}{$W_{\mathbi{T}}^K$} & linear transformation weight matrix of Key\\
&in temporal attention\\
\\
\multirow{2}{*}{$W_{\mathbi{T}}^V$} & linear transformation weight matrix of Value \\
&in temporal attention\\
\\
\multirow{2}{*}{$W_{\mathbi{T}}$} & linear transformation weight matrix of concatenation \\
&in temporal attention\\
\\
\multirow{2}{*}{$E_h$}& attention weight of the $h$-th head in spatiotemporal\\
&fusion attention \\
\\
$A$ & brain effective connectivity matrix\\
\\
$\mathcal{K}$& 3D learnable filter of Fourier attention\\
\\
$\mathcal{X}$ & 3D embedded fMRI data\\
\\
$\mathcal{H}$ & 3D frequency domain fMRI data\\
\\
\multirow{2}{*}{$\tilde{\mathcal{H}}$} & 3D frequency domain fMRI data after applying\\
& learnable filter $\mathcal{K}$\\
\\
$\tilde{\mathcal{X}}$ & 3D denoised fMRI data\\
\\
$\mathcal{Q}_{{\mathbi{T}},h}$ & 3D Query of the $h$-th head in temporal attention\\
\\
$\mathcal{K}_{{\mathbi{T}},h}$ & 3D Key of the $h$-th head in temporal attention\\
\\
$\mathcal{V}_{{\mathbi{T}},h}$ & 3D Value of the $h$-th head in temporal attention\\
\\
\multirow{2}{*}{$\mathcal{Q}_{{\mathbi{S}},h}$} & 3D Query of the $h$-th head in spatiotemporal \\
& fusion attention\\
\\
\multirow{2}{*}{$\mathcal{K}_{{\mathbi{S}},h}$} & 3D Key of the $h$-th head in spatiotemporal \\
& fusion attention\\
\\
\multirow{2}{*}{$\mathcal{O}_{\mathbi{T}}$} & 3D output from the concatenation of multi-head \\
&attention in temporal attention\\
\\
$\mathcal{Z}_{\mathbi{T}}$ & 3D output of temporal attention\\
\\
$\hat{\mathcal{X}}$ & 3D predicted fMRI data\\
\\
$\mathcal{F}(\cdot)$ & 1D fast Fourier transform\\
\\
$\mathcal{F}^{-1}(\cdot)$ & 1D inverse fast Fourier transform\\
\\
$\rho(\cdot)$ & layer normalization operation\\
\\
\multirow{2}{*}{$\Phi(\cdot)$} & a sequence of operations within the feedforward \\
& network module\\
\\
$\gamma(\cdot)$ & concatenation operation\\
\\
$\psi(\cdot)$ &squeeze operation\\
\\
$v_i$& the $i$-th brain region\\
\\
$T$& the number of data points\\
\\
$N$& the number of brain regions\\
\\
$D$& the embedding size of fMRI data\\
\\
$H$& the number of attention heads in multi-head attention\\
\\
$j$& imaginary unit\\
\\\bottomrule
\end{tabular}
\caption{Summary of notations.}
\label{notation}
\end{threeparttable}
\end{table}

\subsection{Proof of Proposition 1}\label{apd_proof}
For a 3D tensor $\mathcal{X}\in\mathbb{R}^{T\times N \times D}$, $\mathcal{X}_{t,n,i}$ denotes the value of the embedded fMRI data $\mathcal{X}$ in a three-dimensional space at a specific time step ($t$), within a particular brain region ($n$), and across a specific dimension ($d$). Let filter $\mathcal{C}_{m,n,i}=\mathcal{F}^{-1}(\mathcal{K}_{f,n,i})$, we have:
\begin{equation}
\mathcal{K}_{f,n,i}=\mathcal{F}(\mathcal{C}_{m,n,i})=\sum_{m=0}^{T-1}\mathcal{C}_{m,n,i}e^{-j \frac{2 \pi}{T} f m}.
\end{equation}

We define cyclic convolution:
\begin{equation}
\tilde{\mathcal{X}}_{t,n,i}=\sum_{m=0}^{T-1}\mathcal{C}_{m,n,i}\mathcal{X}_{(t-m)\, mod\, T,n,i},
\label{cyclic_conv}
\end{equation}
where $mod$ signifies integer modulo operation. Then we have the following equation:
\begin{align*}
\tilde{\mathcal{H}}_{f,n,i}
&=\mathcal{F}(\tilde{\mathcal{X}}_{t,n,i}) \\
&=\sum_{t=0}^{T-1}\tilde{\mathcal{X}}_{t,n,i}e^{-j \frac{2 \pi}{T} f t} \\
&=\sum_{t=0}^{T-1}\sum_{m=0}^{T-1}\mathcal{C}_{m,n,i}\mathcal{X}_{(t-m)\, mod\, T,n,i}e^{-j \frac{2 \pi}{T} f t} \\
&=\sum_{m=0}^{T-1}\mathcal{C}_{m,n,i}e^{-j \frac{2 \pi}{T} f m} \sum_{t=0}^{T-1}\mathcal{X}_{(t-m)\, mod\, T,n,i}e^{-j \frac{2 \pi}{T} f (t-m)} \\
&=\mathcal{K}_{f,n,i}\sum_{t=0}^{T-1}\mathcal{X}_{(t-m)\, mod\, T,n,i}e^{-j \frac{2 \pi}{T} f (t-m)} \\  
&=\mathcal{K}_{f,n,i}(\sum_{t=m}^{T-1}\mathcal{X}_{(t-m)\, mod\, T,n,i}e^{-j \frac{2 \pi}{T} f (t-m)}+\\
&\quad\, \sum_{t=0}^{m-1}\mathcal{X}_{(t-m)\, mod\, T,n,i}e^{-j \frac{2 \pi}{T} f (t-m)}) \\ 
&=\mathcal{K}_{f,n,i}(\sum_{t=0}^{T-m-1}\mathcal{X}_{t,n,i}e^{-j \frac{2 \pi}{T} f t}+\sum_{t=T-m}^{T-1}\mathcal{X}_{t,n,i}e^{-j \frac{2 \pi}{T} f t})\\
&=\mathcal{K}_{f,n,i}\sum_{t=0}^{T-1}\mathcal{X}_{t,n,i}e^{-j \frac{2 \pi}{T} f t}\\
&=\mathcal{K}_{f,n,i}\mathcal{F}(\mathcal{X}_{t,n,i})\\
&=\mathcal{K}_{f,n,i}\mathcal{H}_{f,n,i}.
\end{align*}

Based on the given information, we can conclude that:
\begin{equation}
\tilde{\mathcal{H}}_{f,n,i}=\mathcal{F}(\tilde{\mathcal{X}}_{t,n,i})=\mathcal{K}_{f,n,i}\mathcal{H}_{f,n,i}.
\end{equation}

By definition, the following equation can be derived:
\begin{equation}
\sum_{m=0}^{T-1}\mathcal{C}_{m,n,i}\mathcal{X}_{(t-m)\, mod\, T,n,i}=\tilde{\mathcal{X}}_{t,n,i}=\mathcal{F}^{-1}(\mathcal{K}_{f,n,i}\mathcal{H}_{f,n,i}).
\end{equation}

Rearranging the above equation into the format of a 3D tensor yields:
\begin{equation}
\mathcal{C}\ast \mathcal{X}=\tilde{\mathcal{X}}=\mathcal{F}^{-1}(\mathcal{K}\odot \mathcal{H}),
\end{equation}
where $\ast$ represents cyclic convolution, $\odot$ is the element-wise multiplication. Consequently, we can demonstrate that incorporating a learnable filter $\mathcal{K}$ into the fast Fourier transform and the inverse fast Fourier transform is equivalent to performing a cyclic convolution with the convolution kernel $\mathcal{C}$.

\subsection{Algorithm Description}\label{apd_alg}
The FSTA-EC algorithm consists of two main components: FA and STA. Algorithm 1 provides a summary of FSTA-EC algorithm. FSTA-EC employs an encoder-decoder architecture, with FA functioning as encoder and STA as decoder. The FA-Encoder utilizes FFT to transform the high-noise fMRI data into frequency domain and applies IFFT to restore the denoised fMRI data back to its original physical domain, while concurrently extracting global frequency domain features of the fMRI data throughout this procedure. Additionally, the STA-Decoder captures both the spatiotemporal features of the fMRI data and the spatial effective connectivity across diverse brain regions in a simultaneous manner.

\begin{algorithm}[htb]
    \caption{FSTA-EC}
    \label{alg:algorithm}
    
    \begin{algorithmic}[1] 
    \REQUIRE fMRI time series data.
    \ENSURE Brain effective connectivity $\mathbf{A}$.
        \STATE\textbf{Initialization:} Parameters of FA and STA: $\Psi_f$, $\Psi_{st}$, the training epochs $E$;
        \FOR{$epoch=1\ to\ E$}
        \STATE \textbf{Fourier Attention:} Perform embedding and positional encoding as Eq. (~\ref{PE});
        \STATE Convert fMRI data to the frequency domain with FFT as Eq. (\ref{FFT});
        \STATE Modify the spectrum by applying a learnable filter $\mathcal{K}$ through the modulation process as Eq. (~\ref{filter});
        \STATE Utilize IFFT to map the denoised fMRI data back to the physical domain as Eq.(~\ref{IFFT});
        \STATE \textbf{Spatiotemporal Attention:} Get temporal features $\mathcal{Z}_{\mathbi{T}}$ through temporal attention as Eq.(~\ref{getZ_T});
        \STATE Reweight temporal features $\mathcal{Z}_{\mathbi{T}}$ by brain EC in spatiotemporal fusion attention to acquire spatiotemporal features of fMRI data $\hat{\mathcal{X}}$ as Eq.(~\ref{get_hat_X});
        \ENDFOR
        \STATE Calculate loss and update parameters;
        \RETURN Brain effective connectivity $\mathbf{A}$.
    \end{algorithmic}
\end{algorithm}

\subsection{Evaluation Metrics}\label{apd_metrics}
In order to assess the effectiveness of Fourier spatiotemporal attention (FSTA-EC), we use the following evaluation metrics: Precision, Recall, F1, Accuracy and Structural Hamming distance (SHD). Among them, Recall and Precision are commonly used metrics in brain effective connectivity (EC) learning and other learning tasks. F1 is a harmonic mean that combines Precision and Recall, providing a balanced assessment of both metrics. Accuracy refers to the proportion of samples in the prediction results that are correctly predicted. SHD represents the difference between the learned brain EC and ground-truth brain EC. In general, the higher the Precision, Recall, F1, Accuracy, and the lower the SHD, the better the performance of the method.
The Precision, Recall, F1, Accuracy and SHD can be calculated as follows:

\begin{equation}
Precision=\frac{TP}{TP+FP}
\end{equation}
\begin{equation}
Recall =\frac{TP}{TP+FN}
\end{equation}
\begin{equation}
F1 = \frac{2\times Precision\times Recall}{Precision+ Recall}
\end{equation}
\begin{equation}
Accuracy=\frac{TP+TN}{TP+FP+TN+FN}
\end{equation}
\begin{equation}
SHD=FP+FN
\end{equation}
where $TP$, $FP$, $TN$, and $FN$ denote true positive, false positive, true negative, and false negative, respectively.

\subsection{Additional Experiments}\label{apd_additional_exper}
To further highlight the performance of our method, we incorporated an attention-based approach (MetaRLEC) \cite{MetaRLEC} and a diffusion-based approach (DiffAN) \cite{DiffAN} into our experiments. The results, presented in Table~\ref{result_new}, clearly demonstrate the exceptional performance of FSTA-EC.

\begin{table}[ht]
\centering
\caption{Additional results on simulated datasets.}
\label{result_new} 
\renewcommand\arraystretch{0.8}
\begin{threeparttable}
\setlength{\tabcolsep}{1.5mm}{ 
\begin{tabular}{r| r| c c|c }
\toprule \multirow{3}{*}{Data}& \multirow{3}{*}{Metrics}  &
\multicolumn{3}{c}{Methods}\\\cmidrule{3-5}
  &   & DiffAN & MetaRLEC & FSTA-EC\\
  &  & \citeyear{DiffAN} & \citeyear{MetaRLEC} & (\textbf{Ours}) \\ \midrule
\multirow{5}{*}{Sim1}
   &  Precision $\uparrow$ &  0.54$\pm$0.22 & \underline{0.67$\pm$0.08} & \textbf{0.78\boldmath $\pm$0.11} \\
   &  Recall $\uparrow$ &  \underline{0.56$\pm$0.17} & 0.35$\pm$0.12 & \textbf{0.78\boldmath $\pm$0.11} \\
   & F1 $\uparrow$ & \underline{0.55$\pm$0.12} & 0.46$\pm$0.06 & \textbf{0.77\boldmath $\pm$0.09} \\
   & Accuracy $\uparrow$ & 0.77$\pm$0.10 & \underline{0.81$\pm$0.06} & \textbf{0.89\boldmath $\pm$0.04} \\
   & SHD $\downarrow$ & 5.78$\pm$2.53 & \underline{5.13$\pm$0.83} & \textbf{2.70\boldmath $\pm$1.00} \\
\midrule
\multirow{5}{*}{Sim2}
    &  Precision $\uparrow$ &  0.54$\pm$0.18 & \underline{0.81$\pm$0.06} & \textbf{0.85\boldmath $\pm$0.15} \\ 
    &  Recall $\uparrow$ & 0.51$\pm$0.11 & \underline{0.57$\pm$0.13} & \textbf{0.87\boldmath $\pm$0.09} \\
    & F1 $\uparrow$ & 0.52$\pm$0.13 & \underline{0.67$\pm$0.10} & \textbf{0.85\boldmath $\pm$0.11} \\
    & Accuracy $\uparrow$ & 0.73$\pm$0.09 & \underline{0.84$\pm$0.05} & \textbf{0.91\boldmath $\pm$0.08} \\
    & SHD $\downarrow$ & 6.75$\pm$2.18 & \underline{4.03$\pm$1.63} & \textbf{2.20\boldmath $\pm$1.94} \\
 \midrule
\multirow{5}{*}{Sim3}
   &  Precision $\uparrow$ & 0.57$\pm$0.14 & \underline{0.93$\pm$0.14} & \textbf{0.96\boldmath $\pm$0.09} \\
   &  Recall $\uparrow$ & 0.52$\pm$0.14 & \textbf{0.64\boldmath $\pm$0.08} & \underline{0.61$\pm$0.07} \\
   & F1 $\uparrow$ & 0.54$\pm$0.11 & \textbf{0.76\boldmath $\pm$0.09} &  \underline{0.74$\pm$0.06} \\
   & Accuracy $\uparrow$ & 0.75$\pm$0.16 & \underline{0.84$\pm$0.03} & \textbf{0.88\boldmath $\pm$0.03} \\
   & SHD $\downarrow$ & 6.33$\pm$2.43 & \underline{2.98$\pm$0.82} & \textbf{2.95\boldmath $\pm$0.74} \\
\midrule
\multirow{5}{*}{Sim4}
    &  Precision $\uparrow$ & 0.51$\pm$0.10 & \textbf{0.94\boldmath $\pm$0.08} & \underline{0.75$\pm$0.13} \\
    &  Recall $\uparrow$ & 0.29$\pm$0.05 & \underline{0.40$\pm$0.03} & \textbf{0.59\boldmath $\pm$0.03} \\
    & F1 $\uparrow$ & 0.37$\pm$0.05 & \underline{0.56$\pm$0.05} & \textbf{0.66\boldmath $\pm$0.06} \\
    & Accuracy $\uparrow$ & 0.81$\pm$0.02 & \underline{0.88$\pm$0.05} & \textbf{0.88\boldmath $\pm$0.03} \\
    & SHD $\downarrow$ & 18.95$\pm$2.00 & \underline{12.30$\pm$1.97} & \textbf{11.85\boldmath $\pm$2.76} \\
 \bottomrule
\end{tabular} } 
\end{threeparttable} 
\end{table}

\end{document}